\documentclass[ijoc,sglanonrev,letterpaper,onecolumn]{informs4}

\OneAndAHalfSpacedXII
\usepackage[letterpaper,margin=1in]{geometry}
\usepackage{natbib}
\bibpunct[, ]{(}{)}{,}{a}{}{,}%
\def\bibfont{\small\setlength{\baselineskip}{14.5pt}}%

\usepackage{fix-cm}
\usepackage{anyfontsize}
\usepackage{makecell, rotating, float, array, booktabs, amsmath, amssymb, mathtools, mathrsfs, multirow, tabularx, ltablex, bm, graphicx, tikz, etoolbox, placeins}
\usepackage{xcolor}
\usepackage{listings}
\definecolor{codebg}{RGB}{247,249,252}
\definecolor{codeframe}{RGB}{183,193,207}
\definecolor{codekeyword}{RGB}{0,75,150}
\definecolor{codeapi}{RGB}{112,62,152}
\definecolor{codecomment}{RGB}{45,112,72}
\definecolor{codestring}{RGB}{168,52,58}
\definecolor{codeconstant}{RGB}{178,91,0}
\lstdefinestyle{torchdcmPython}{
    linewidth=\linewidth,
    language=Python,
    basicstyle=\scriptsize\ttfamily,
    keywordstyle=\color{codekeyword}\bfseries,
    commentstyle=\color{codecomment}\itshape,
    stringstyle=\color{codestring},
    morekeywords=[2]{torch,masked_fill,full,scatter_reduce,exp,zeros_like,scatter_add,log,unsqueeze,sum_by_unit,logsumexp,as_tensor},
    keywordstyle=[2]\color{codeapi}\bfseries,
    morekeywords=[3]{True,False,None,inf,float64,dtype,device,dim,reduce,include_self},
    keywordstyle=[3]\color{codeconstant}\bfseries,
    backgroundcolor=\color{codebg},
    rulecolor=\color{codeframe},
    columns=flexible,
    keepspaces=true,
    showstringspaces=false,
    breaklines=true,
    breakatwhitespace=true,
    tabsize=2,
    numbers=none,
    captionpos=t,
    aboveskip=0pt,
    belowskip=0pt,
    lineskip=-1pt,
    frame=single,
    framesep=5pt,
    xleftmargin=0pt,
    xrightmargin=0pt,
}
\makeatletter
\renewcommand{\lst@makecaption}[2]{%
    \@maketablecaption{#1}{#2}\vskip5pt%
}
\makeatother
\usepackage[linesnumbered,ruled,vlined]{algorithm2e}
\SetAlFnt{\scriptsize}
\SetNlSty{scriptsize}{}{}
\SetAlgoNlRelativeSize{0}
\SetAlCapFnt{\EGT\TableNameFontStyle}
\SetAlCapNameFnt{\EGT\TableCaptionFontStyle}
\SetAlgoSkip{medskip}
\usepackage[T1]{fontenc}
\usepackage[final]{microtype}
\keepXColumns
\usepackage{longtable}
\usepackage{hyperref}
\hypersetup{
    colorlinks,
    linkcolor={red!50!red},
    citecolor={blue!50!blue},
    urlcolor={blue!80!blue},
    pdfinfo={
        Title={TorchDCM: A Unified PyTorch-Native Package for Discrete Choice Modeling},
        Author={Baichuan Mo, Zhengzhong Ricky You, Xiqun Michael Chen, Ruimin Li},
        Subject={Discrete choice modeling software},
        Keywords={discrete choice; PyTorch; econometric inference; GPU; software}
    }
}
\usepackage{enumitem}
\usepackage{needspace}
\usepackage{xspace}
\setlist[itemize]{label=\textbullet,topsep=4pt,itemsep=2pt,parsep=0pt,partopsep=0pt}
\newcommand{\torchdcm}{\textsc{TorchDCM}\xspace}
\newcommand{\cmark}{\ensuremath{\checkmark}}
\graphicspath{{./figures/}}

\TheoremsNumberedThrough
\EquationsNumberedThrough

\begin{document}

\RUNAUTHOR{Mo et al.}
\RUNTITLE{TorchDCM: A Unified PyTorch-Native Package}

\TITLE{TorchDCM: A Unified PyTorch-Native\\ Package for Discrete Choice Modeling}

\ARTICLEAUTHORS{%
    \AUTHOR{%
        Baichuan Mo\textsuperscript{1},
        Zhengzhong Ricky You\textsuperscript{1,*},
        Xiqun Michael Chen\textsuperscript{2},
        Ruimin Li\textsuperscript{1}
    }%
    \AFF{%
        \textsuperscript{1}Department of Civil Engineering, Tsinghua University,
        Beijing 100084, China, \EMAIL{bmo@tsinghua.edu.cn}\\
        \textsuperscript{2}Institute of Intelligent Transportation Systems,
        College of Civil Engineering and Architecture, Zhejiang University,
        Hangzhou, China, \EMAIL{chenxiqun@zju.edu.cn}\\
        \textsuperscript{*}Corresponding author: \EMAIL{ricky.you.or@gmail.com}
    }%
}

\ABSTRACT{Estimating large and simulation-intensive discrete choice models (DCMs) requires repeated evaluation of utilities, probabilities, derivatives, and simulated likelihoods over many observations, alternatives, and draws. Existing DCM software provides mature econometric workflows, while recent GPU-oriented tools accelerate selected models, leaving a gap between econometric coverage and scalable differentiable computation. We introduce \torchdcm, an open Python package for discrete choice modeling that compiles choice data and model specifications into a unified PyTorch-native likelihood engine for estimation, inference, prediction, and structured reporting on CPU or CUDA devices. The package covers the principal econometric functionality available across Biogeme and Apollo, including multinomial, nested, mixed, ordered, latent-variable, and panel likelihoods. It also supports ragged choice sets, constrained parameters, covariance estimation, willingness-to-pay analysis, elasticities, and extensible likelihood components. We evaluate \torchdcm against seven other estimation packages in aligned synthetic and real-data full-estimation experiments. \torchdcm completes all 45 synthetic cases, runs fastest in every comparable synthetic case, and satisfies the prespecified final-log-likelihood tolerance in every comparison with at least two comparable solutions. More precisely, it reduces median runtime by 89.1\%--99.7\% relative to Biogeme and Apollo across model-data settings. CUDA provides an additional 12.0--71.0$\times$ speedup over single-core \torchdcm. These results establish a scalable and reproducible foundation for econometric estimation and differentiable choice-model development. The open-source package and executed examples are available at \url{https://github.com/mbc96325/torchdcm}.}

\KEYWORDS{discrete choice modeling; PyTorch; GPU computing; econometric estimation; open-source software.}

\maketitle

\section{Introduction}
\label{sec:introduction}

Discrete choice models (DCMs) provide an important and widely used framework for analyzing decisions among a finite set of alternatives, with applications in transportation, marketing, operations, and public policies \citep{mcfadden1974conditional,benakiva1985discrete,train2009discrete}. Estimating modern DCMs requires repeated evaluation of utilities, probabilities, derivatives, and simulated likelihoods across large sets of observations, alternatives, and simulation draws. These calculations benefit from vectorized tensor operations, automatic differentiation, and accelerator execution. Econometric practice simultaneously requires choice-specific data structures, constrained estimation, covariance analysis, and post-estimation tools.

We introduce \torchdcm, an open-source Python package for discrete choice modeling that brings these computational and econometric requirements together in a unified PyTorch-native likelihood engine. The package compiles choice data and model specifications into reusable tensors and provides a consistent workflow for estimation, inference, prediction, simulation, and reporting across CPUs and CUDA-enabled GPUs. Its extensible model interface supports multinomial, nested, mixed, ordered, latent-variable, and panel likelihoods.

The existing DCM software ecosystem is valuable but built around different computational abstractions. Biogeme combines symbolic model specification with compiled numerical backends \citep{bierlaire2023biogeme}, Apollo provides a flexible choice-modeling framework in R \citep{hess2019apollo}, and \texttt{mlogit} and \texttt{gmnl} provide established estimators for random utility models and preference heterogeneity \citep{croissant2020mlogit,sarrias2017gmnl}. As the numbers of observations, alternatives, parameters, or simulation draws increase, model construction, compilation, interpreter, and data-conversion overheads may become material because these econometric interfaces are not organized around a unified PyTorch tensor and device-execution model. Conversely, PyTorch supplies efficient tensor operations, automatic differentiation, and accelerator execution \citep{paszke2019pytorch}, but it does not provide choice-specific data structures, constrained estimation, econometric inference, or post-estimation tools.

Differences among these abstractions also create translation and validation costs. Moving the same specification between packages requires reconciling wide and long data, availability rules, alternative coding, fixed-parameter conventions, scale normalizations, panel identifiers, simulation draws, and covariance definitions. Differences in Hessian, robust covariance, willingness-to-pay (WTP), elasticity, and probability outputs further complicate numerical auditing. Researchers extending a likelihood or comparing estimators may therefore need to rebuild data, optimization, inference, and reporting workflows.

\torchdcm combines the econometric capabilities expected from traditional DCM interfaces with PyTorch-native tensor computation and device execution, thereby addressing these gaps. Through a unified CPU/CUDA workflow, it provides discrete-choice-native data and specification objects, ragged and panel tensors, constrained estimation, automatic differentiation, covariance calculation, prediction, WTP, elasticities, and structured reports.

We separately develop a benchmark and validation pipeline that compares \torchdcm with \texttt{torch-choice}, Biogeme, Apollo, SciPy, \texttt{mlogit}, \texttt{gmnl}, and \texttt{xlogit} \citep{du2023torchchoice,bierlaire2023biogeme,hess2019apollo,virtanen2020scipy,croissant2020mlogit,sarrias2017gmnl,arteaga2022xlogit}. The pipeline aligns data, specifications, starting values, runtime scope, and numerical diagnostics across estimators. It generates synthetic datasets that vary sample size, number of alternatives, number of variables, feature correlation, random-coefficient dimension, and simulation draws. It also collects public datasets from transportation, product and brand choice, energy, recreation, and service-choice applications. The reported experiments cover multinomial logit (MNL), nested logit (NL), mixed logit (MixL), ordered logit, ordered probit, latent-class, hybrid-choice, and panel models.

This paper makes three contributions.
\begin{itemize}
    \item \textit{We develop a unified PyTorch-native package for discrete choice modeling.} \torchdcm connects choice-data construction, utility specification, maximum-likelihood estimation, econometric inference, prediction, WTP, elasticities, and structured reporting through a unified interface spanning the supported model families.

    \item \textit{We design an extensible tensor engine for econometric estimation and inference.} The engine combines ragged choice sets, availability masks, panel aggregation, seeded simulation, constrained parameter transformations, automatic differentiation, and a unified CPU/CUDA execution path. New likelihood components can therefore reuse the package's data, optimization, covariance, prediction, and reporting services.

    \item \textit{We conduct an aligned benchmark of DCM estimation software.} The benchmark compares eight packages on configurable synthetic cases and public datasets under common specifications, starting values, runtime scopes, and numerical diagnostics. It reports final-likelihood agreement, runtime, failures, and stress-test limits.
\end{itemize}

Across the aligned experiments, \torchdcm matches comparable final log likelihoods, completes every synthetic stress case, and substantially reduces estimation time relative to Biogeme and Apollo. CUDA execution provides further gains for large and simulation-intensive cases. Section~\ref{sec:results} reports the complete experimental design and results.

The remainder of the paper proceeds as follows. Section~\ref{sec:related} reviews the existing choice-modeling software landscape and positions \torchdcm within it. Section~\ref{sec:overview} introduces notation for DCMs and summarizes likelihood computation. Section~\ref{sec:design} presents the package architecture and its user-facing objects. Section~\ref{sec:implementation} explains how the package implements likelihood evaluation, simulation, and inference. Section~\ref{sec:results} introduces the benchmarking framework and reports results on numerical parity and computational scaling. Finally, Section~\ref{sec:conclusion} summarizes the main findings, discusses the limitations, and outlines directions for future work.

\section{Related Software}
\label{sec:related}

The software landscape spans domain-specific econometric packages and general numerical or GPU-oriented tools. In this section, we review these two groups separately to show how each supports a different part of the DCM estimation workflow.

\subsection{Econometric choice-modeling packages}

Biogeme provides a mature Python environment for specifying and estimating a broad range of DCMs, while Apollo offers a similarly broad and flexible modeling environment in R \citep{bierlaire2023biogeme,hess2019apollo}. Their main use case is conventional econometric choice analysis that requires established model syntax, estimation, and reporting. We therefore use them as reference implementations in this study. However, these workflows can become computationally costly as data and simulation dimensions grow, particularly when repeated likelihood evaluations do not use a unified accelerator-execution path.

The R packages \texttt{mlogit} and \texttt{gmnl} provide established estimators for random utility models and preference heterogeneity \citep{croissant2020mlogit,sarrias2017gmnl}. Their main use case is formula-based estimation of MNL and related extensions within the R statistical ecosystem, rather than a general interface spanning multiple DCM likelihood families.

\subsection{General-purpose numerical and GPU-oriented software}

SciPy supplies general-purpose optimization and scientific-computing routines \citep{virtanen2020scipy}. In DCM estimation, it optimizes a likelihood constructed separately by the researcher, leaving data representation, probability construction, inference, and reporting to the application.

\texttt{xlogit} is a Python choice-modeling package that emphasizes computational performance and provides GPU-accelerated MixL estimation \citep{arteaga2022xlogit}. It is particularly useful when one of its supported models can benefit from dedicated accelerator execution, but it does not aim to reproduce the breadth of a general econometric choice-modeling environment.

\texttt{torch-choice} is an open-source PyTorch package for large-scale conditional logit and NL estimation \citep{du2023torchchoice}. It provides a memory-efficient \texttt{ChoiceDataset}, formula- and dictionary-based model specification, availability sets, regularized estimation, prediction, and CPU/GPU execution. The two packages overlap on PyTorch-native MNL and NL estimation, while \torchdcm also supports MixL, ordered, latent-class, hybrid-choice, error-component, and simulated panel likelihoods, together with WTP, elasticities, covariance estimation, and reporting.

PyTorch provides automatic differentiation, flexible tensor operations, and CPU/GPU execution for general differentiable computation \citep{paszke2019pytorch}. It supports custom numerical models, but does not by itself supply choice-specific data, likelihood, inference, or reporting interfaces.

\subsection{Summary and positioning of TorchDCM}

The software reviewed above addresses complementary aspects of DCM estimation. Biogeme and Apollo provide comprehensive environments for conventional model specification, estimation, and reporting, whereas \texttt{mlogit} and \texttt{gmnl} offer more focused, formula-based workflows for MNL and related extensions. \texttt{torch-choice} brings conditional logit and NL estimation into a PyTorch-native interface. SciPy, \texttt{xlogit}, and PyTorch serve distinct computational roles, supporting generic optimization, accelerated estimation for selected choice models, and general-purpose tensor computation, respectively. More broadly, stable domain-specific representations and model transformations can connect expressive modeling interfaces to reusable numerical services \citep{legat2021mathoptinterface,chen2023rsome,besancon2023diffopt}.

At the intersection of these econometric and computational tools, \torchdcm organizes its supported model families around a unified application programming interface, tensor representation, and result object. Ragged and panel choice data share this representation, while the separate benchmark supports consistent comparisons across estimation packages. By exposing likelihood components within a unified PyTorch-native workflow, \torchdcm lets researchers develop differentiable choice models while reusing data handling, model-specific parameter transformations and optimization, covariance estimation, prediction, and reporting. Its separation of user-facing model structure from performance-critical numerical operations follows established computational software designs \citep{rodriguez2023pynumero,wouda2024pyvrp}.

\begin{table}[htb]
\caption{Model and feature coverage summary for DCM software.}
\label{tab:software_support}
\centering
\scriptsize
\renewcommand{\arraystretch}{0.96}
\begin{tabular*}{\linewidth}{@{\extracolsep{\fill}}lccccccc@{}}
\toprule
Capability & \torchdcm & \texttt{torch-choice} & Biogeme & Apollo & \texttt{mlogit} & \texttt{gmnl} & \texttt{xlogit} \\
\midrule
MNL & \cmark & \cmark & \cmark & \cmark & \cmark & \cmark & \cmark \\
NL & \cmark & \cmark & \cmark & \cmark & \cmark & -- & -- \\
CNL & \cmark & -- & \cmark & \cmark & -- & -- & -- \\
MixL & \cmark & -- & \cmark & \cmark & \cmark & \cmark & \cmark \\
WTP-space MixL & \cmark & -- & \cmark & \cmark & -- & -- & \cmark \\
Scale heterogeneity & \cmark & -- & \cmark & \cmark & \cmark & \cmark & -- \\
Latent class & \cmark & -- & \cmark & \cmark & -- & \cmark & -- \\
Ordered logit/probit & \cmark & -- & \cmark & \cmark & -- & -- & -- \\
Hybrid choice & \cmark & -- & \cmark & \cmark & -- & -- & -- \\
Named error components & \cmark & -- & \cmark & \cmark & -- & -- & -- \\
Simulated panel likelihood & \cmark & -- & \cmark & \cmark & \cmark & \cmark & \cmark \\
Alternative availability masks & \cmark & \cmark & \cmark & \cmark & \cmark & \cmark & \cmark \\
Ragged long format & \cmark & -- & -- & -- & \cmark & -- & -- \\
Documented GPU estimation & \cmark & \cmark & -- & -- & -- & -- & \cmark \\
\bottomrule
\end{tabular*}
\par\vspace{5pt}
\parbox{\linewidth}{\footnotesize\textit{Notes.} CNL: cross-nested logit. Ragged long format denotes a direct interface for unequal-length choice sets rather than conversion to a dense availability tensor. \cmark: documented package interface or official example. --: no documented direct interface.}
\end{table}

We summarize documented model, data, and device coverage in Table~\ref{tab:software_support} to clarify the position of \torchdcm within the choice-modeling literature. The distinction is the combination of capabilities. Among the packages summarized, \torchdcm pairs the broadest econometric model coverage with ragged long-format data and documented GPU estimation in one unified PyTorch-native interface. \texttt{torch-choice} provides a PyTorch-native CPU/GPU workflow for conditional logit and NL, while the remaining packages contribute complementary econometric or computational capabilities. None combines the full set within the same extensible CPU/CUDA workflow.

\section{Overview of DCM Estimation}
\label{sec:overview}

In this section, we summarize the notation, data layouts, and likelihood computations needed to follow the package design and implementation.

Let \(\mathcal{N}\) denote the set of decision makers. Consider first an individual \(n\in\mathcal{N}\) observed in one choice situation. The individual faces a feasible choice set \(\mathcal{C}_n\) and chooses \(y_n\in\mathcal{C}_n\). The set \(\mathcal{C}=\bigcup_{n\in\mathcal{N}}\mathcal{C}_n\) contains all alternatives represented in the data. Let \(\mathcal{K}\) denote the set of free utility coefficients, with \(\boldsymbol{\beta}\in\mathbb{R}^{\lvert\mathcal{K}\rvert}\). A random utility model writes the utility of alternative \(i\in\mathcal{C}_n\) as \(U_{n,i}=V_{n,i}(\boldsymbol{\beta})+\varepsilon_{n,i}\), where \(V_{n,i}(\boldsymbol{\beta})\) is the systematic utility, \(\boldsymbol{\beta}\) is a vector of taste coefficients, and \(\varepsilon_{n,i}\) is an unobserved component. For example, the familiar linear specification is \(V_{n,i}(\boldsymbol{\beta})=\boldsymbol{x}_{n,i}^{\mathsf T}\boldsymbol{\beta}\), where \(\boldsymbol{x}_{n,i}\) collects alternative attributes and alternative-specific intercept indicators. The choice probability is then \(P_{n,i}(\boldsymbol{\beta})=\Pr\left(U_{n,i}\ge U_{n,j},\ \forall j\in\mathcal{C}_n\right)\). Under independently and identically distributed type-I extreme-value errors, this probability reduces to the MNL probability \(P_{n,i}(\boldsymbol{\beta})=\exp\{V_{n,i}(\boldsymbol{\beta})\}/\sum_{j\in\mathcal{C}_n}\exp\{V_{n,j}(\boldsymbol{\beta})\}\) \citep{mcfadden1974conditional,train2009discrete}.

Some datasets contain repeated choices from the same individual. Let \(\mathcal{P}_n\) denote the set of observed choice occasions for \(n\in\mathcal{N}\), and denote the observation at occasion \(u\in\mathcal{P}_n\) by \(n_u\). The complete set of choice observations is \(\mathcal{O}=\{n_u:n\in\mathcal{N},\ u\in\mathcal{P}_n\}\). Each observation \(n_u\in\mathcal{O}\) has a feasible set \(\mathcal{C}_{n_u}\), a chosen alternative \(y_{n_u}\), and an optional weight \(w_{n_u}\). Cross-sectional data are the special case in which each \(\mathcal{P}_n\) contains one occasion. For fixed-coefficient models, the weighted log likelihood is therefore
\begin{align}
\ell(\boldsymbol{\beta})
&=\sum_{n_u\in\mathcal{O}}
w_{n_u}\log P_{n_u,y_{n_u}}(\boldsymbol{\beta})
=\sum_{n\in\mathcal{N}}\sum_{u\in\mathcal{P}_n}
w_{n_u}\log P_{n_u,y_{n_u}}(\boldsymbol{\beta}).
\label{eq:mnl_likelihood}
\end{align}
Eq.~\eqref{eq:mnl_likelihood} already requires stable masked normalization over every observed feasible choice set.

Choice data are commonly stored in wide or long format. A wide table uses one row per choice observation and stores alternative-specific variables in separate column blocks. A long table instead uses one row for each feasible observation--alternative pair. We denote the set of long-format rows by \(\mathcal{M}=\{(n_u,i):n_u\in\mathcal{O},\ i\in\mathcal{C}_{n_u}\}\), which contains \(\lvert\mathcal{M}\rvert=\sum_{n_u\in\mathcal{O}} \lvert\mathcal{C}_{n_u}\rvert\) rows. We use \(m\in\mathcal{M}\) to index an alternative row. This representation allows observations to have different choice-set sizes without padding them to a common width.

The model can be generalized in several directions. NL and cross-nested logit (CNL) replace the MNL assumption that the $\varepsilon_{n,i}$ are independent with a generalized extreme-value error structure that induces dependence among alternatives in a common nest. CNL permits an alternative to be associated with more than one nest. MixL replaces fixed $\boldsymbol{\beta}$ with a random coefficient distribution, so estimation adds its mean vector and dispersion parameters (standard deviations or a Cholesky covariance factor) and integrates the choice probabilities over that distribution \citep{hensher2003mixed,train2009discrete}. Hybrid choice models further add parameters for the latent-variable structural equation, its links to utility, measurement loadings, and measurement-error variances \citep{walker2002generalized}. Panel data additionally require individual-level likelihood aggregation. Each generalization requires repeated, numerically stable evaluations of masked probabilities, group reductions, simulated likelihoods, gradients, and Hessians.

\section{Package Design}
\label{sec:design}

A \torchdcm estimation begins with two independent inputs. A data object represents observations, alternatives, availability, weights, and panel structure, while a specification object defines named parameters and systematic utility. The selected model compiles these inputs once, passes the resulting tensors through its likelihood kernel, and uses the unified PyTorch-native engine for derivative evaluation and optimization. When estimation ends, the fitted result retains the same model and data objects for inference, prediction, and reporting.

\begin{figure}[htb]
    \centering
    \includegraphics[width=\linewidth]{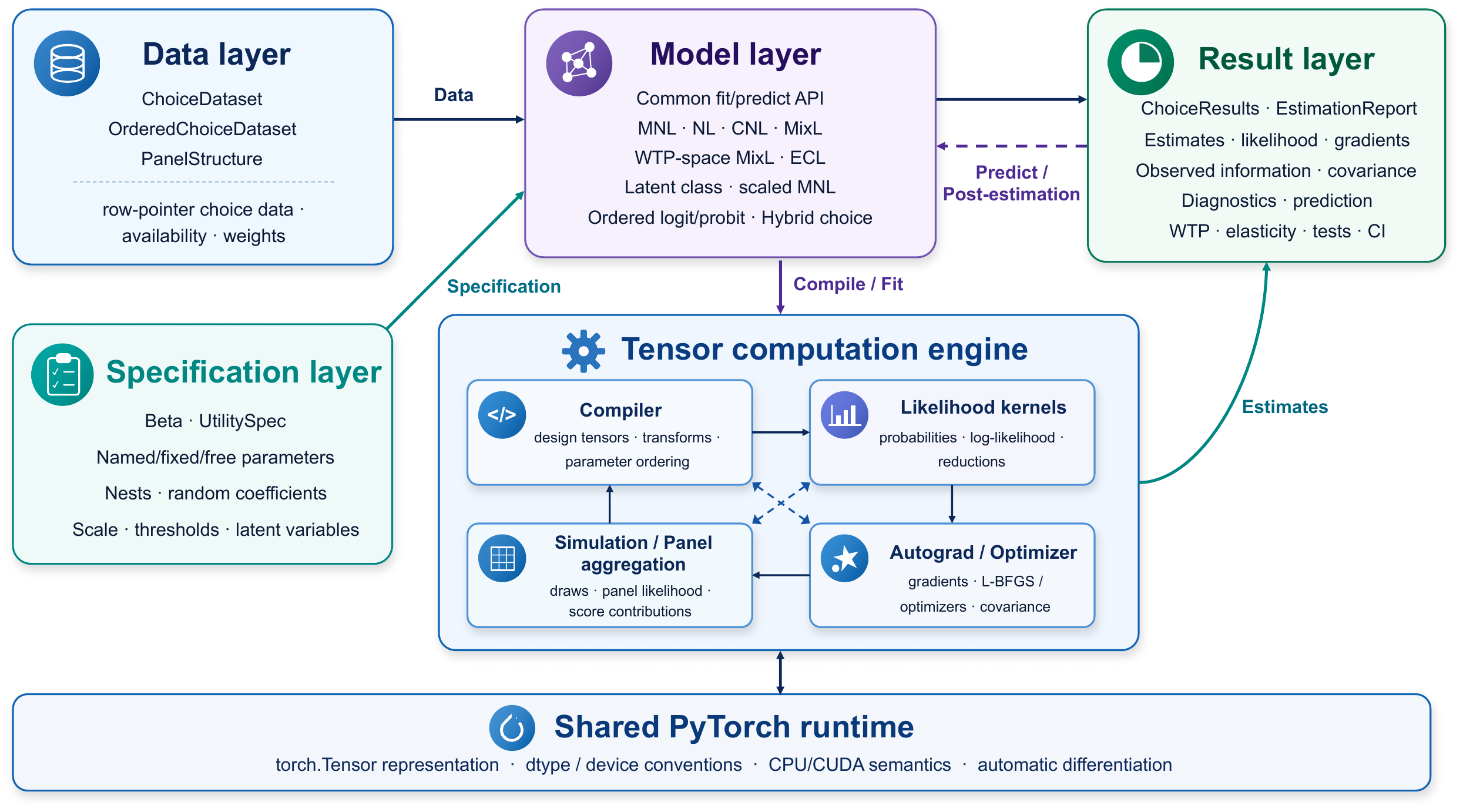}
    \caption{TorchDCM package architecture. Data and specification objects feed
    a unified model layer, which dispatches compilation, likelihood evaluation,
    simulation, panel aggregation, automatic differentiation, and optimization
    to the tensor engine. Result objects retain the fitted model and data for
    prediction and post-estimation.}
    \label{fig:torchdcm_framework}
\end{figure}

Figure~\ref{fig:torchdcm_framework} traces this workflow across the package architecture. The data and specification layers meet only at the model layer, allowing either input to be constructed and checked independently. A call to \texttt{fit} then connects model compilation, likelihood evaluation, simulation or panel aggregation when required, automatic differentiation, and optimization. All stages follow unified PyTorch tensor, dtype, and device conventions, and the result layer reuses the compiled task rather than reconstructing it for each post-estimation operation.

\subsection{Installation}

\torchdcm is distributed through the \href{https://pypi.org/project/torchdcm/}{Python Package Index}. Its source code and executed examples are available in the \href{https://github.com/mbc96325/torchdcm}{public GitHub repository}. The package supports Python 3.10 or later and can be installed together with its required dependencies using a single command:

\begin{center}
\fcolorbox{codeframe}{codebg}{%
    \parbox{0.38\linewidth}{\centering\texttt{pip install torchdcm}}}
\end{center}

\Needspace{5\baselineskip}
\subsection{Data layer}

Choice datasets vary in storage format, feasible-set size, availability, weighting, and panel structure. Padding every observation to the largest choice-set width wastes storage and arithmetic when feasible sets differ. \texttt{ChoiceDataset} instead stores feasible observation--alternative rows contiguously; \texttt{obs\_ptr} stores the row-slice boundaries for each choice observation. Alternative identifiers, chosen rows, variables, availability indicators, and weights remain aligned with this row order, which supports segmented reductions for both balanced and ragged data.

The \texttt{ChoiceDataset} class constructs this representation from a pandas long table through \texttt{from\_long}. Its \texttt{from\_wide} method instead accepts alternative-specific column mappings. Listing~\ref{lst:data_construction} shows both interfaces. \texttt{long\_df} contains one row per feasible observation--alternative pair. \texttt{wide\_df} contains one row per observation and columns such as \texttt{TRAIN\_time}, \texttt{BUS\_time}, and \texttt{CAR\_time}. In \texttt{from\_wide}, the template \texttt{\{"time": "\{alt\}\_time"\}} expands the column name once for each alternative. Users may instead supply an explicit alternative-to-column mapping. Availability, weights, observation variables, individual identifiers, dtype, and device can be supplied through optional arguments.

\begin{lstlisting}[caption={Constructing long- and wide-format choice datasets.},label={lst:data_construction}]
from torchdcm import ChoiceDataset

# long_df columns: choice_id, alternative, chosen, time, cost
long_data = ChoiceDataset.from_long(long_df, obs_id="choice_id",
    alt_id="alternative", choice="chosen", variables=["time", "cost"])

# wide_df includes TRAIN, BUS, and CAR time and cost columns
# "{alt}_time" expands once for each alternative
wide_data = ChoiceDataset.from_wide(wide_df, alternatives=["TRAIN", "BUS", "CAR"],
    choice="chosen_alt", variables={"time": "{alt}_time", "cost": "{alt}_cost"})
data = wide_data
\end{lstlisting}

Construction checks the required columns, verifies exactly one chosen row per observation, and requires the chosen alternative to be available. It also checks that weights, individual identifiers, and observation-level variables are constant within each observation and that row- and observation-level tensors have compatible lengths. Optional individual identifiers create a \texttt{PanelStructure}, while ordered responses use \texttt{OrderedChoiceDataset}. The resulting data object is reused by compilation, likelihood and panel aggregation, covariance construction, prediction, and reporting.

\subsection{Specification layer}

The specification input must translate an econometric utility equation into a form that can be compiled without losing parameter names, starting values, or identification restrictions. \texttt{UtilitySpec} stores one symbolic expression for each alternative, and each \texttt{Beta} stores a parameter name, initial value, and fixed/free status. This construction separates the systematic utility definition from both the source-data layout and the selected probability model.

Listing~\ref{lst:utility_specification} shows the named-expression interface and its equivalent linear-formula interface.

\begin{lstlisting}[caption={Defining systematic utilities through two equivalent interfaces.},label={lst:utility_specification}]
from torchdcm import Beta, UtilitySpec

# Named expressions: V_CAR = B_ASC + B_TIME*time + B_COST*cost
B_ASC, B_TIME, B_COST = Beta("B_ASC"), Beta("B_TIME"), Beta("B_COST")
spec = UtilitySpec()
spec.utility("CAR", B_ASC + B_TIME * "time" + B_COST * "cost")
spec.utility("BUS", B_TIME * "time" + B_COST * "cost")
spec.utility("TRAIN", B_TIME * "time" + B_COST * "cost")

# Equivalent formulas; uppercase names are parameters
spec = UtilitySpec.from_formula({
    "CAR": "B_ASC + B_TIME * time + B_COST * cost",
    "BUS": "B_TIME * time + B_COST * cost",
    "TRAIN": "B_TIME * time + B_COST * cost",
})
\end{lstlisting}

Model-specific objects add nests, random coefficients, WTP coefficients, latent classes, scale equations, thresholds, or latent-variable components without changing the base utility specification. The model compiler reuses the parameter names, restrictions, and expressions to create transformations and a stable ordering for likelihood evaluation, optimization, covariance calculation, prediction, and result tables.

\subsection{Model layer}

The model layer must bind the independent data and specification inputs to the probability structure that defines the likelihood. Each model class therefore selects a likelihood kernel and validates any additional structure, such as nests, random coefficients, scale equations, or latent variables, before optimization begins. The available classes cover MNL, NL, CNL, MixL, WTP-space MixL, error-components logit, latent-class logit, alternative-specific and covariate-scaled MNL, ordered logit and probit, and hybrid-choice models.

Listing~\ref{lst:model_configuration} illustrates model-specific structure and numerical configuration for NL and MixL.

\begin{lstlisting}[caption={Configuring NL and MixL estimators.},label={lst:model_configuration}]
import torch
from torchdcm import MixedLogit, NestedLogit, RandomCoefficient

# NL: nests, precision, device, and optimizer limit
nests = {"PUBLIC": ["TRAIN", "BUS"], "PRIVATE": ["CAR"]}
nl = NestedLogit(spec, nests, dtype=torch.float64, device="cuda", max_iter=300)
nl_result = nl.fit(data, cov_type="robust")

# MixL: random coefficient and reproducible simulation draws
random = [RandomCoefficient("B_TIME", sigma_init=0.2)]
mixl = MixedLogit(spec, random, n_draws=512, seed=12345, device="cuda", max_iter=300)
mixl_result = mixl.fit(data)
\end{lstlisting}

Each constructor selects and validates the corresponding likelihood structure. The unified \texttt{fit(data)} interface compiles the inputs, dispatches likelihood and derivative evaluation and optimization to the tensor engine, and returns the same result interface across model families. Supported models also accept run-specific covariance and cluster options.

\subsection{Tensor computation engine}

Estimation repeatedly evaluates the same data-dependent likelihood structure at different parameter values. Reconstructing expressions and indices at every optimization step would therefore incur unnecessary computational overhead. The tensor computation engine avoids this overhead by separating one-time model construction from repeated numerical evaluation. During compilation, the engine constructs the design tensors and grouping indices, applies parameter transformations, and establishes a consistent parameter ordering. Given the current parameter values and cached tensors, the selected kernel then computes utilities, probabilities, and observation-level log-likelihood contributions. The engine invokes the relevant simulation or panel routines only when the model requires draws, latent quantities, or aggregation over repeated choices. Automatic differentiation in PyTorch provides the derivatives used by the optimization and covariance estimation routines.

Users need not configure a separate engine. Dataset and model \texttt{dtype}/\texttt{device} settings, simulation arguments, and \texttt{fit} options determine compilation and execution. For example, \texttt{model.fit(data, max\_iter=300)} triggers one-time compilation followed by repeated likelihood and derivative evaluation. The cached representation is reused for likelihood, gradient, Hessian, score, prediction, and covariance calculations. Section~\ref{sec:implementation} details this process.

\subsection{Result layer}

Applied estimation requires reusable econometric outputs rather than an isolated coefficient tensor. When optimization terminates, \texttt{ChoiceResults} retains the fitted model and data together with parameter estimates, likelihood values, derivatives, covariance matrices, and convergence diagnostics. It derives standard errors, confidence intervals, predictions, WTP, and elasticities from this retained state when the corresponding operation is defined.

Listing~\ref{lst:result_reuse} illustrates inference, prediction, and reporting from a fitted result.

\begin{lstlisting}[caption={Reusing fitted results.},label={lst:result_reuse}]
# Select an available covariance
result = nl_result
classic_cov = result.cov_params("classic")
robust_result = result.get_robustcov_results("robust")

# Predict with the retained model and data
probabilities = robust_result.predict_proba()
choices = robust_result.predict()

# Render and export the structured estimation report
report = robust_result.report(confidence_level=0.95)
print(report.to_text())
files = robust_result.save_report("output", formats=("html", "json", "csv", "latex"))
\end{lstlisting}

The report includes model, data, fit, inference, parameter, and warning
information. These calls reuse the retained estimation task. In the electronic
companion (EC), we reproduce the rendered report in Section EC.1.

\subsection{Unified PyTorch-native runtime}

The workflow uses a unified \texttt{torch.Tensor} representation from input compilation through result reuse. Dataset transfer and model construction follow the same \texttt{dtype} and \texttt{device} conventions, so the utility specification and estimation call remain unchanged when execution moves between CPU and CUDA. The default \texttt{float64} dtype supports econometric estimation, and automatic differentiation in PyTorch provides a unified derivative interface across the supported model families. Listing~\ref{lst:device_selection} selects CUDA when available and otherwise uses CPU.

\begin{lstlisting}[caption={Selecting CPU or CUDA execution.},label={lst:device_selection}]
import torch
from torchdcm import MixedLogit

device = torch.device("cuda" if torch.cuda.is_available() else "cpu")
model = MixedLogit(spec, random, n_draws=256, device=device)
result = model.fit(data)
\end{lstlisting}

The \texttt{fit} call transfers the dataset and creates the design tensors, grouping structures, and simulation draws on the selected device. These tensors remain there throughout likelihood evaluation, differentiation, and covariance calculation. CUDA is most useful when large data or draw dimensions offset kernel-launch overhead, while small models may remain faster on CPU.

\subsection{Customization and extension}

A custom likelihood component must return one differentiable contribution for each choice observation. The public extension point \texttt{MultinomialLogit.loglike\_per\_obs(params, data, compiled=None)} receives the current free-parameter tensor, a \texttt{ChoiceDataset}, and the cached \texttt{CompiledUtility}. Its output is a tensor of length \texttt{data.n\_obs}, whose sum defines the objective passed to the package optimizer.

Listing~\ref{lst:custom_likelihood} adds a ridge penalty to the MNL likelihood and distributes the penalty equally across observations. Constructing the subclass in place of \texttt{MultinomialLogit} requires no separate registration call. The inherited \texttt{fit} method then reuses compilation, automatic differentiation, optimization, classic or robust covariance calculation, prediction, and reporting. Models that also change the probability rule can additionally override \texttt{predict\_proba}.

\begin{lstlisting}[caption={Extending the observation-level MNL likelihood.},label={lst:custom_likelihood}]
from torchdcm import MultinomialLogit

class RidgeMNL(MultinomialLogit):
    def loglike_per_obs(self, params, data, compiled=None):
        base = super().loglike_per_obs(params, data, compiled)
        penalty = 1e-4 * params.square().sum() / data.n_obs
        return base - penalty

result = RidgeMNL(spec).fit(data, cov_type="robust")
\end{lstlisting}

\section{Implementation}
\label{sec:implementation}

In this section, we explain how \torchdcm evaluates compiled likelihoods and derivatives within a unified PyTorch-native tensor workflow. We then show how the same workflow supports covariance estimation, simulation, panel likelihoods, and advanced model components.

\subsection{Compiled tensor likelihoods}

At every optimizer iteration, the feasible alternative rows must be normalized within each choice observation, even when choice-set sizes vary. Rebuilding the utility expressions and grouping indices at each parameter update would add unnecessary overhead to this repeated computation. \torchdcm therefore compiles the utility representation once and uses segmented tensor reductions for ragged choice sets.

On the first likelihood call for a data--model pair, the utility expression is materialized as a free-parameter design tensor \(\boldsymbol{X}\) and, when needed, a fixed-parameter tensor \(\boldsymbol{X}^{\text{F}}\). Utility evaluation in subsequent optimizer iterations is reduced to \(\boldsymbol{V}=\boldsymbol{X}\boldsymbol{\beta}+\boldsymbol{X}^{\text{F}}\boldsymbol{\beta}^{\text{F}}\). The tensors, parameter ordering, chosen-row indices, and grouping indices are cached and reused by likelihood, gradient, prediction, and Hessian calls. This avoids repeatedly parsing utility expressions or reconstructing data-dependent indices inside the optimization loop.

The likelihood kernel selects its normalization operation from the choice-set layout. With a common choice-set width, utilities are reshaped to an observation-by-alternative tensor and normalized along the alternative dimension. With unequal widths, the row-to-observation index derived from \texttt{obs\_ptr} defines tensor segments. A segment maximum is first subtracted for numerical stability, after which \texttt{scatter\_add} computes one denominator per observation. Availability is imposed before these reductions, so each denominator includes only feasible alternatives.

Listing~\ref{lst:ragged_reduction} shows the latter kernel. All rows are reduced in a small number of device-native tensor operations, with no observation-level Python loop.

\begin{lstlisting}[caption={GPU-native ragged-choice likelihood reduction (abridged).},label={lst:ragged_reduction}]
masked_u = utility.masked_fill(~data.availability, -torch.inf)
max_u = torch.full(
    (data.n_obs,), -torch.inf,
    dtype=utility.dtype, device=utility.device,
)
max_u = max_u.scatter_reduce(
    0, row_to_obs, masked_u, reduce="amax", include_self=True,
)
exp_u = torch.exp(masked_u - max_u[row_to_obs])
exp_u = exp_u.masked_fill(~data.availability, 0.0)
denom = torch.zeros_like(max_u).scatter_add(
    0, row_to_obs, exp_u,
)
chosen_u = utility[data.chosen_row]
return data.weights * (chosen_u - max_u - torch.log(denom))
\end{lstlisting}

\subsection{Automatic differentiation and inference}

Econometric estimation must enforce parameter restrictions during optimization
and report uncertainty on the natural parameter scale. \torchdcm handles both
through one automatic-differentiation path. Automatic differentiation has long
provided accurate and efficient derivatives for large-scale optimization
without separate hand-coded gradient routines \citep{bischof1997automatic}.
PyTorch's limited-memory
Broyden--Fletcher--Goldfarb--Shanno (L-BFGS) optimizer
\citep{liu1989limited,paszke2019pytorch} operates on an unrestricted vector
\(\boldsymbol{\phi}\), while each likelihood evaluation applies
\(\boldsymbol{\theta}=T(\boldsymbol{\phi})\). The transformation is the identity
for unrestricted coefficients, uses logistic or softplus maps for bounded or
positive parameters, and constructs ordered thresholds from cumulative
positive increments.

After estimation, the parameter covariance matrix supplies standard errors,
hypothesis tests, and confidence intervals. Classic maximum-likelihood
inference uses the inverse observed information. Because optimization occurs
on the internal parameter scale, \torchdcm maps this inverse information to
the reported scale through the Jacobian of \(T(\cdot)\). Robust and cluster
covariance estimators use the same transformed inverse-information matrix as
the sandwich bread and place observation- or cluster-level score outer
products in the middle.

At the solution, define
\(\boldsymbol{H}_{\phi}=-\nabla^2_{\boldsymbol{\phi}}
\ell(T(\boldsymbol{\phi}))\) and
\(\boldsymbol{J}_{T}=\partial T/\partial\boldsymbol{\phi}^{\mathsf T}\).
The classic covariance is
\(\boldsymbol{\Sigma}_{\mathrm{classic}}
=\boldsymbol{J}_{T}\boldsymbol{H}_{\phi}^{+}
\boldsymbol{J}_{T}^{\mathsf T}\), where \((\cdot)^{+}\) denotes the
pseudoinverse. Let
\(\boldsymbol{s}_{o}=\nabla_{\boldsymbol{\theta}}
\ell_o(\widehat{\boldsymbol{\theta}})\) be the score for observation
\(o\in\mathcal{O}\). For a set of clusters \(\mathcal{G}\), define
\(\boldsymbol{s}_{g}=\sum_{o\in g}\boldsymbol{s}_{o}\) for
\(g\in\mathcal{G}\). The robust and cluster covariance estimators are
\begin{align}
\boldsymbol{\Sigma}_{\mathrm{robust}}
 &=\boldsymbol{\Sigma}_{\mathrm{classic}}
\left(\sum_{o\in\mathcal{O}}
\boldsymbol{s}_{o}\boldsymbol{s}_{o}^{\mathsf T}\right)
\boldsymbol{\Sigma}_{\mathrm{classic}},
\qquad
\boldsymbol{\Sigma}_{\mathrm{cluster}}
 =\boldsymbol{\Sigma}_{\mathrm{classic}}
\left(\sum_{g\in\mathcal{G}}
\boldsymbol{s}_{g}\boldsymbol{s}_{g}^{\mathsf T}\right)
\boldsymbol{\Sigma}_{\mathrm{classic}}.
\label{eq:covariance_estimators}
\end{align}
The robust covariance estimator in Eq.~\eqref{eq:covariance_estimators} allows misspecification of the observation-level likelihood. The cluster estimator additionally allows dependence within decision makers, households, markets, or other known groups. Both are currently available for MNL and NL. Simulation-based and other extended models report classic covariance.

For balanced MNL, \torchdcm evaluates the closed-form scores for all
observations at once, stacks them into \(\boldsymbol{S}\), and computes the
robust middle matrix as the device-native Gram product
\(\boldsymbol{S}^{\mathsf T}\boldsymbol{S}\). Cluster scores are first
aggregated by group and then use the same product. Other supported likelihoods
obtain scores by differentiating their observation-level contributions. The
transformation Jacobian and covariance operations remain on the selected
device.

Algorithm~\ref{alg:unified_fit} summarizes this parameter-scale handling.

\begin{algorithm}[H]
\caption{Constrained estimation and covariance transformation}
\label{alg:unified_fit}
\DontPrintSemicolon
\KwIn{Compiled task $\mathscr{Q}$, likelihood $\ell(T(\boldsymbol{\phi});\mathscr{Q})$, transformation $T$, and initial internal parameters}
\KwOut{Natural-scale estimates $\widehat{\boldsymbol{\theta}}$ and covariance $\boldsymbol{\Sigma}_{\mathrm{classic}}$}
Initialize unconstrained parameters $\boldsymbol{\phi}$\;
\While{the stopping criterion is not satisfied}{
    $\boldsymbol{\theta} \leftarrow T(\boldsymbol{\phi})$\tcp*{map to the natural parameter scale}
    $\ell \leftarrow \operatorname{LogLike}(\boldsymbol{\theta};\mathscr{Q})$\;
    $\boldsymbol{g}_{\phi} \leftarrow \nabla_{\boldsymbol{\phi}}[-\ell]$ by automatic differentiation\;
    Update $\boldsymbol{\phi}$ with PyTorch L-BFGS\;
}
$\widehat{\boldsymbol{\theta}} \leftarrow T(\widehat{\boldsymbol{\phi}})$\;
$\boldsymbol{H}_{\phi}\leftarrow-\nabla^2_{\boldsymbol{\phi}}\ell(T(\widehat{\boldsymbol{\phi}});\mathscr{Q})$\;
$\boldsymbol{J}_{T}\leftarrow \partial T/\partial\boldsymbol{\phi}^{\mathsf T}$ at $\widehat{\boldsymbol{\phi}}$\;
$\boldsymbol{\Sigma}_{\mathrm{classic}}\leftarrow
\boldsymbol{J}_{T}\boldsymbol{H}_{\phi}^{+}\boldsymbol{J}_{T}^{\mathsf T}$\;
\Return{$\widehat{\boldsymbol{\theta}},\boldsymbol{\Sigma}_{\mathrm{classic}}$}\;
\end{algorithm}

\subsection{Simulation and panel likelihoods}

Simulation-based models repeatedly evaluate probabilities conditional on random
coefficients or latent variables. \torchdcm compiles a fixed draw matrix once
and evaluates all draws in batches, keeping the objective deterministic across
optimizer iterations.

For an observation \(o\in\mathcal{O}\) with latent vector
\(\boldsymbol{z}\), let \(P_o(\boldsymbol{\theta},\boldsymbol{z})\) denote the
conditional probability of the observed choice. Its simulated probability is
\begin{align}
P_o(\boldsymbol{\theta})
&=\int P_o(\boldsymbol{\theta},\boldsymbol{z})
 f(\boldsymbol{z})\,d\boldsymbol{z}
\approx \widehat{P}_o(\boldsymbol{\theta})
=\frac{1}{R}\sum_{r=1}^{R}
P_o(\boldsymbol{\theta},\boldsymbol{z}_r).
\label{eq:simulated_probability}
\end{align}
Let \(\mathcal{D}\) index the dimensions of the latent vector. The matrix
\(\boldsymbol{Z}\in\mathbb{R}^{R\times\lvert\mathcal{D}\rvert}\) stacks
the \(R\) integration points \(\boldsymbol{z}_r\) in Eq.~\eqref{eq:simulated_probability} over dimensions \(\mathcal{D}\). The integration
points can be generated from a fixed seed, supplied by the user, and optionally
paired antithetically. For correlated random coefficients, let
\(\boldsymbol{\Omega}\) denote the covariance matrix and
\(\boldsymbol{L}\) its Cholesky factor. Then
\(\boldsymbol{Z}\boldsymbol{L}^{\mathsf T}\) produces correlated deviations,
with \(\boldsymbol{\Omega}=\boldsymbol{L}\boldsymbol{L}^{\mathsf T}\).

For MixL, let \(\mathcal{K}^{\mathrm{R}}\subseteq\mathcal{K}\) index the random
coefficients. A full calculation stores one complete coefficient vector per
draw in \(\boldsymbol{B}\in
\mathbb{R}^{R\times\lvert\mathcal{K}\rvert}\) and evaluates
\(\boldsymbol{U}_{\mathrm{full}}=\boldsymbol{X}\boldsymbol{B}^{\mathsf T}
+\left(\boldsymbol{X}^{\text{F}}\boldsymbol{\beta}^{\text{F}}\right)
\boldsymbol{1}_{R}^{\mathsf T}\), where \(\boldsymbol{X}\) is the design matrix,
\(\boldsymbol{X}^{\text{F}}\boldsymbol{\beta}^{\text{F}}\) is the fixed
utility contribution, and \(\boldsymbol{1}_{R}\) is an \(R\)-vector of ones.
Because only the coefficients indexed by \(\mathcal{K}^{\mathrm{R}}\) vary
across draws, let \(\boldsymbol{\mu}\) be the full mean vector and let
\(\boldsymbol{B}^{\text{R}}\), \(\boldsymbol{X}^{\text{R}}\), and
\(\boldsymbol{\mu}^{\text{R}}\) denote the coefficient, design, and mean blocks
restricted to these indices. \torchdcm computes the mean utility once and adds
only the draw-specific deviations:
\begin{align}
\boldsymbol{U}
&=\left(\boldsymbol{X}\boldsymbol{\mu}
+\boldsymbol{X}^{\text{F}}\boldsymbol{\beta}^{\text{F}}\right)
 \boldsymbol{1}_{R}^{\mathsf T}
+\boldsymbol{X}^{\text{R}}
\left(\boldsymbol{B}^{\text{R}}
-\boldsymbol{1}_{R}(\boldsymbol{\mu}^{\text{R}})^{\mathsf T}\right)^{\mathsf T}.
\label{eq:mixl_decomposition}
\end{align}
Comparing this full calculation with the decomposition in Eq.~\eqref{eq:mixl_decomposition}, the cost falls from
\(O(\lvert\mathcal{M}\rvert R\lvert\mathcal{K}\rvert)\) to
\(O(\lvert\mathcal{M}\rvert\lvert\mathcal{K}\rvert+
\lvert\mathcal{M}\rvert R\lvert\mathcal{K}^{\mathrm{R}}\rvert)\), with the
largest gain when
\(\lvert\mathcal{K}^{\mathrm{R}}\rvert\ll\lvert\mathcal{K}\rvert\).

For panel data, all occasions from one decision maker must share the same
draw. Let \(n_u\) identify occasion \(u\in\mathcal{P}_n\). Then
\begin{align}
A_{n,r}
&=\sum_{u\in\mathcal{P}_n}\log P_{n_u,y_{n_u}\mid r},
&& n\in\mathcal{N},\quad r=1,\ldots,R,
\label{eq:panel_draw_total}\\
\ell(\boldsymbol{\theta})
&=\sum_{n\in\mathcal{N}}\left[
\operatorname{logsumexp}_{r=1,\ldots,R}(A_{n,r})
-\log R\right].
\label{eq:panel_simulated_likelihood}
\end{align}
\torchdcm forms the draw totals in Eq.~\eqref{eq:panel_draw_total} with one
\texttt{index\_add} operation and applies the \texttt{logsumexp} operation in
Eq.~\eqref{eq:panel_simulated_likelihood} only after occasion-level probabilities have been combined.
Listing~\ref{lst:mixl_panel} shows both the MixL update and panel reduction.

\Needspace{0.22\textheight}
\begin{lstlisting}[caption={Random-block MixL utility update and vectorized panel integration (condensed).},label={lst:mixl_panel}]
# Update only the random-coefficient columns across draws.
utility = (design @ means).unsqueeze(1)
delta = transformed_betas[:, free_mask] - means[free_idx].unsqueeze(0)
utility = utility + design[:, free_idx] @ delta.T

# Sum repeated choices by individual, then integrate over draws.
unit_log_prob = panel.sum_by_unit(obs_log_prob)  # index_add on axis 0
panel_ll = torch.logsumexp(unit_log_prob, dim=1)
panel_ll = panel_ll - torch.log(
    torch.as_tensor(n_draws, dtype=utility.dtype, device=utility.device)
)
\end{lstlisting}

\subsection{Advanced model components}

Advanced models change utility, dependence, class membership, or measurement
components while reusing the same normalization, simulation, differentiation,
and reporting services.

For the WTP-space MixL specification, \(\alpha\) denotes the cost coefficient,
\(c_m\) the cost in row \(m\), and \(x_{m,k}\) the value of attribute \(k\)
in that row. The set \(\mathcal{K}^{\mathrm{WTP}}\) indexes attributes with
draw-specific WTP coefficients \(\omega_{k,r}\). Let \(V_m^{(0)}\) denote the
remaining deterministic component of utility, yielding
\(U_{m,r}=V_m^{(0)}+\alpha\bigl(c_m+
\sum_{k\in\mathcal{K}^{\mathrm{WTP}}}\omega_{k,r}\cdot x_{m,k}\bigr)\).
The compiler constructs only the matrix
\(\boldsymbol{W}\in\mathbb{R}^{R\times
\lvert\mathcal{K}^{\mathrm{WTP}}\rvert}\), where \(W_{r,k}=\omega_{k,r}\),
and reuses the existing MixL simulation and panel reductions.

Error-components logit (ECL) introduces shared shocks that correlate alternatives.
Let \(V_m\) denote systematic utility excluding these components. For component
set \(\mathcal{H}\), loading \(\lambda_{m,h}\), scale
\(\sigma_h\), and standardized draw \(z_{h,r}\), utility becomes
\(U_{m,r}=V_m+\sum_{h\in\mathcal{H}}\lambda_{m,h}\cdot\sigma_h\cdot z_{h,r}\).
The loadings are compiled as row-level tensors and the components as zero-mean
random coefficients, so ECL uses the existing MixL kernel rather than a
separate likelihood.

For a latent-class model, \(\mathcal{S}\) is the class set,
\(\pi_{n,s}\) is the membership probability, and
\(P_{n,y_n\mid s}\) is the class-conditional choice probability, giving
\(\ell_n=\operatorname{logsumexp}_{s\in\mathcal{S}}
\bigl(\log\pi_{n,s}+\log P_{n,y_n\mid s}\bigr)\).
One membership logit is fixed for identification. The same classwise
log-joint tensor supplies the likelihood and posterior class probabilities.

Hybrid choice models augment the conditional choice likelihood with structural
latent variables and indicator measurement densities. For observation
\(o\in\mathcal{O}\), draw \(r\), and latent dimension
\(l\in\mathcal{L}\), let \(\mu_{o,l}\) denote the structural mean,
\(\sigma_l\) the latent-variable scale, and \(z_{r,l}\) a standardized draw.
The kernel first evaluates
\(\eta_{o,r,l}=\mu_{o,l}+\sigma_l\cdot z_{r,l}\), then adds the corresponding
choice and measurement contributions and integrates over the draws using
Eq.~\eqref{eq:panel_simulated_likelihood}. Prediction uses posterior draw
weights derived from the same measurement densities.

\section{Benchmark Design and Results}
\label{sec:results}

In this section, we define the benchmark scope and data and then report the synthetic, device, and actual-data results. A snapshot containing the code, data, configurations, and replicated results for all reported experiments is available in the \href{https://github.com/mbc96325/torchdcm-evaluation-benchmark#experiment-to-output-mapping}{\texttt{torchdcm-evaluation-benchmark}} repository.

\subsection{Benchmark design and data}

Following the package-level comparison in Table~\ref{tab:software_support}, we compare \torchdcm with Torch-Choice, Biogeme, Apollo, \texttt{mlogit}, \texttt{gmnl}, and \texttt{xlogit} when an aligned specification and executable wrapper are available. SciPy provides a generic optimization baseline for selected MNL specifications. We therefore compare only the applicable subset of packages in each model--data setting.

The design follows established guidance for controlled computational experiments and testbeds by aligning problem factors, runtime scope, randomization, and comparison criteria \citep{mcgeoch1996experimental,eckman2023simopt}.

All cross-estimator experiments run on an AMD Ryzen 9 9950X3D CPU with 16 physical cores and 32 threads. For equal-resource comparison, each estimator process and its child processes are pinned to one logical CPU. OpenMP and Basic Linear Algebra Subprograms (BLAS) thread environment variables are set to one, PyTorch intra- and inter-operation thread counts are set to one, and Apollo uses \texttt{nCores=1}. TorchDCM and Torch-Choice both use \texttt{float64} tensors. Their MNL and NL runs receive one untimed forward--backward warm-up before optimization, so the one-time PyTorch kernel and automatic-differentiation initialization is excluded for both packages. Reported runtime is parameter estimation plus covariance construction. Data generation or loading, aligned-design construction, interpreter and package startup, and file input/output are excluded. Compilation performed inside an estimator call remains included. The 300-second stress limit is applied to the complete worker wall clock. Each dataset--estimator case runs in a fresh process to prevent compiler state or memory reuse across rows. The device-isolation experiments in Section~\ref{sec:device_results} compare one logical CPU with an NVIDIA GeForce RTX 5090 with 32 GB of memory. All runtimes in the result tables are reported in seconds. Within each row, estimators use aligned data, model specifications, and starting values. Simulated models also use common draws where supported.

For each result row, let \(\mathcal{B}_{\mathrm{ok}}\) contain the estimators that return a finite, completed solution and define the best final log likelihood as \(\ell^\star=\max_{b\in\mathcal{B}_{\mathrm{ok}}}\ell_b\). Let \(N_{\mathrm{obs}}=\lvert\mathcal{O}\rvert\) denote the number of choice observations in that row. A dagger marks a completed solver for which \(\ell^\star-\ell_b\) exceeds \(\tau_{\ell}=\max\{0.25,10^{-5}|\ell^{\star}|,0.01N_{\mathrm{obs}}\}\). A ``Yes'' in the agreement column means that at least two completed, non-daggered estimators remain and their maximum pairwise final-likelihood difference does not exceed \(\tau_\ell\). A ``Fail'' denotes an attempted run that does not complete successfully, for example because of optimizer failure, non-finite diagnostics, or a singular covariance calculation. ``Timeout'' identifies a stress worker that reaches the 300-second limit or is externally terminated under the same constrained run, and ``N.A.'' denotes fewer than two comparable estimates.

\textit{Synthetic data.} Let \(\mathcal{N}\), \(\mathcal{C}\), and \(\mathcal{K}\) denote the sets of choice situations, alternatives, and observed variables in a synthetic case. The controlled generator varies their cardinalities \(\lvert\mathcal{N}\rvert\), \(\lvert\mathcal{C}\rvert\), and \(\lvert\mathcal{K}\rvert\), together with equicorrelation \(\rho\). The equicorrelation matrix \(\boldsymbol{\Sigma}_{\rho}\in \mathbb{R}^{\lvert\mathcal{K}\rvert\times\lvert\mathcal{K}\rvert}\) has \((\boldsymbol{\Sigma}_{\rho})_{k,k}=1\) and \((\boldsymbol{\Sigma}_{\rho})_{k,k'}=\rho\) for \(k\neq k'\). For each \(j\in\mathcal{C}\), it draws $\boldsymbol{x}_{n,j}\sim\mathcal{N}(\boldsymbol{0},\boldsymbol{\Sigma}_{\rho})$ and forms $V_{n,j}=\alpha_j+\boldsymbol{x}_{n,j}^{\mathsf T}\boldsymbol{\beta}$. MNL choices are sampled from the resulting MNL probabilities using a fixed seed. For every NL case, choices are instead regenerated from a two-level NL data-generating process. The two non-singleton nests use dissimilarity parameters \(\lambda_{\mathrm{A}}=0.65\) and \(\lambda_{\mathrm{B}}=0.80\). A singleton nest is fixed at one. For every MixL case, the first \(\lvert\mathcal{K}^{\mathrm{R}}\rvert\) coefficients are independently generated as \(\beta_{n,k}=\beta_k+\sigma_k z_{n,k}\), where \(z_{n,k}\sim\mathcal{N}(0,1)\), and choices are sampled from the resulting conditional MNL probabilities. The true \(\sigma_k\) values are evenly spaced from 0.4 to 0.2. The remaining coefficients are fixed. The full design varies one dimension at a time, and the stress design reaches \(\lvert\mathcal{N}\rvert=50{,}000\), \(\lvert\mathcal{C}\rvert=35\), and \(\lvert\mathcal{K}\rvert=20\). MixL estimation uses the values of \(\lvert\mathcal{K}^{\mathrm{R}}\rvert\) reported in the tables and \(R=32\) shared antithetic draws for TorchDCM and Biogeme. Both estimators use the common starting value \(\sigma_k=0.3\) and the smooth positive-scale parameterization \(\sigma_k=\operatorname{softplus}(\phi_k)\). Apollo uses 32 Halton draws and is included for runtime comparison. This construction separates computational scaling in \(\lvert\mathcal{N}\rvert\), \(\lvert\mathcal{C}\rvert\), \(\lvert\mathcal{K}\rvert\), \(\rho\), and \(\lvert\mathcal{K}^{\mathrm{R}}\rvert\) from differences in empirical data preparation.

\textit{Actual data.} The empirical benchmark combines public Biogeme case-study data, datasets distributed with the R \texttt{mlogit} package, the London Passenger Mode Choice (LPMC) data, and a processed mode-choice sample from the 2022 National Household Travel Survey (NHTS) \citep{bierlaire2023biogeme,croissant2020mlogit,hillel2019lpmc,fhwa2022nhts}. Table~\ref{tab:actual_data} reports the exact data entering estimation after case-specific filters. The table reports \(\lvert\mathcal{N}\rvert\) choice situations, \(\lvert\mathcal{M}\rvert\) alternative rows processed by the likelihood, \(\lvert\mathcal{C}\rvert\) distinct alternatives or ordered response levels, and \(\lvert\mathcal{K}\rvert\) free parameters in the aligned MNL specification. For Optima, \(\mathcal{K}\) indexes the aligned ordered-response parameters. ModeCanada and RiskyTransport have ragged choice sets, so \(\lvert\mathcal{M}\rvert<\lvert\mathcal{N}\rvert \lvert\mathcal{C}\rvert\). All other unordered cases use balanced row layouts and availability masks as needed.

\begin{table}[htb]
\caption{Actual datasets used in the benchmark.}
\label{tab:actual_data}
\centering
\scriptsize
\renewcommand{\arraystretch}{0.96}
\begin{tabular*}{\linewidth}{@{\extracolsep{\fill}}llrrrrll@{}}
\toprule
Data & Application & $|\mathcal{N}|$ & $|\mathcal{M}|$ & $|\mathcal{C}|$ & $|\mathcal{K}|$ & Layout & Models evaluated \\
\midrule
Swissmetro & Intercity mode & 10,719 & 32,157 & 3 & 4 & Balanced & MNL, NL, MixL \\
NHTS 2022 & Trip mode & 27,375 & 136,875 & 5 & 16 & Balanced & MNL, NL \\
Airline itinerary & Itinerary choice & 3,609 & 10,827 & 3 & 5 & Balanced & MNL, NL, MixL \\
Parking Spain & Parking facility & 1,576 & 4,728 & 3 & 5 & Balanced & MNL, NL, MixL \\
Telephone & Service plan & 434 & 2,170 & 5 & 5 & Balanced & MNL, MixL \\
LPMC London & Urban mode & 81,086 & 324,344 & 4 & 5 & Balanced & MNL, NL, MixL \\
Catsup & Brand choice & 2,798 & 11,192 & 4 & 3 & Balanced & MNL, NL, MixL \\
Cracker & Brand choice & 3,292 & 13,168 & 4 & 3 & Balanced & MNL, NL, MixL \\
Electricity & Contract choice & 4,308 & 17,232 & 4 & 6 & Balanced & MNL, NL, MixL \\
HC & Heating/cooling system & 250 & 1,750 & 7 & 2 & Balanced & MNL, NL, MixL \\
Heating & Heating system & 900 & 4,500 & 5 & 2 & Balanced & MNL, NL, MixL \\
Mode & Travel mode & 453 & 1,812 & 4 & 2 & Balanced & MNL, NL, MixL \\
NOx & Control technology & 632 & 9,480 & 15 & 3 & Balanced & MNL, MixL \\
RiskyTransport & Risky transport mode & 1,793 & 5,405 & 4 & 7 & Ragged & MNL, MixL \\
Train & Rail service & 2,929 & 5,858 & 2 & 4 & Balanced & MNL, MixL \\
Fishing & Fishing mode & 1,182 & 4,728 & 4 & 5 & Balanced & MNL, NL, MixL \\
ModeCanada & Intercity mode & 4,324 & 15,520 & 4 & 3 & Ragged & MNL, MixL \\
Optima & Ordered attitude response & 1,822 & 1,822 & 6 & 9 & Ordered & Ordered logit/probit \\
\bottomrule
\end{tabular*}
\end{table}

Validation experiments for ordered logit, ordered probit, latent-class,
hybrid-choice, and panel models are reported in Sections EC.2 and EC.3.

\subsection{Synthetic-data results}

The synthetic experiments first vary \(\lvert\mathcal{N}\rvert\), \(\lvert\mathcal{C}\rvert\), \(\lvert\mathcal{K}\rvert\), and \(\rho\) in MNL, NL, and MixL specifications. All estimators use the same generated long-format data. The MNL and NL coefficients start at zero. The MixL mean coefficients also start at zero, while each random-coefficient scale starts at 0.3. The MNL benchmark includes all eight estimators. The NL benchmark reports TorchDCM, Torch-Choice, Biogeme, and Apollo, while the MixL benchmark reports TorchDCM, Biogeme, and Apollo because the current pipeline provides aligned implementations under the common draw and parameterization protocol for these packages.

\begin{table}[htb]
\caption{Synthetic controlled MNL runtime comparison (seconds).}
\label{tab:synthetic_controlled}
\centering
\scriptsize
\renewcommand{\arraystretch}{0.96}
\setlength{\tabcolsep}{1.3pt}
\begin{tabular*}{\linewidth}{@{\extracolsep{\fill}}ll*{12}{r}l@{}}
\toprule
Sweep & Case & $|\mathcal{N}|$ & $|\mathcal{C}|$ & $|\mathcal{K}|$ & $\rho$
& TorchDCM & Torch-Choice & SciPy & Biogeme & Apollo & \texttt{mlogit} & \texttt{gmnl}
& \texttt{xlogit} & Consistent? \\
\midrule
Sample size & $|\mathcal{N}|=1{,}000$ & 1000 & 4 & 6 & 0.30
& 0.005 & 0.006 & 0.761 & 2.231 & 0.353 & 0.032 & 0.118 & 0.010 & Yes \\
Sample size & $|\mathcal{N}|=10{,}000$ & 10000 & 4 & 6 & 0.30
& 0.015 & 0.028 & 8.954 & 2.214 & 1.029 & 0.334 & 1.833 & 0.074 & Yes \\
Sample size & $|\mathcal{N}|=100{,}000$ & 100000 & 4 & 6 & 0.30
& 0.107 & 0.229 & 106.261 & 3.520 & 20.587 & 2.805 & 13.466 & 0.734 & Yes \\
Alternatives & $|\mathcal{C}|=3$ & 20000 & 3 & 6 & 0.30
& 0.024 & 0.036 & 23.971 & 1.762 & 1.855 & 0.495 & 2.039 & 0.098 & Yes \\
Alternatives & $|\mathcal{C}|=10$ & 20000 & 10 & 6 & 0.30
& 0.086 & 0.223 & 25.729 & 6.883 & 5.965 & 1.624 & 17.134 & 0.722 & Yes \\
Alternatives & $|\mathcal{C}|=20$ & 20000 & 20 & 6 & 0.30
& 0.337 & 2.633 & 21.923 & 22.980 & 23.805 & 3.415 & 78.993 & 4.060 & Yes \\
Variables & $|\mathcal{K}|=4$ & 20000 & 5 & 4 & 0.30
& 0.029 & 0.058 & 10.722 & 2.060 & 1.914 & 0.768 & 3.018 & 0.137 & Yes \\
Variables & $|\mathcal{K}|=12$ & 20000 & 5 & 12 & 0.30
& 0.044 & 0.085 & 32.483 & 5.904 & 5.163 & 0.944 & 14.772 & 0.460 & Yes \\
Variables & $|\mathcal{K}|=32$ & 20000 & 5 & 32 & 0.30
& 0.087 & 0.193 & 40.246 & 41.282 & 27.004 & 1.563 & 80.933 & 2.880 & Yes \\
Correlation & $\rho=0.00$ & 20000 & 5 & 12 & 0.00
& 0.043 & 0.086 & 27.920 & 5.418 & 4.808 & 0.923 & 11.751 & 0.455 & Yes \\
Correlation & $\rho=0.50$ & 20000 & 5 & 12 & 0.50
& 0.041 & 0.080 & 44.127 & 5.880 & 5.140 & 1.023 & 14.412 & 0.478 & Yes \\
Correlation & $\rho=0.98$ & 20000 & 5 & 12 & 0.98
& 0.053 & 0.106 & 20.545 & 6.109 & 4.897 & 1.030 & 15.981 & 0.474 & Yes \\
Stress & Small & 30000 & 20 & 12 & 0.50
& 0.723 & 5.589 & 43.010 & 75.347 & 61.989 & 6.013 & 201.058 & 10.694 & Yes \\
Stress & Medium & 40000 & 28 & 16 & 0.50
& 2.480 & 17.085 & 73.944 & Timeout & 236.970 & 13.672 & Timeout & 43.686 & Yes \\
Stress & Large & 50000 & 35 & 20 & 0.50
& 5.203 & 35.282 & 101.333 & Timeout & Timeout & 25.269 & Timeout & 127.409 & Yes \\
\bottomrule
\end{tabular*}
\end{table}

Table~\ref{tab:synthetic_controlled} shows that TorchDCM has the lowest runtime in every controlled MNL case, while every completed estimator satisfies the final-likelihood criterion. Torch-Choice completes all 15 cases and ranks second in 13. An implementation difference in the benchmarked versions helps explain this gap. TorchDCM compiles utility expressions once into a contiguous row-by-parameter design, so each optimizer evaluation uses a direct matrix product and log-sum-exp reduction. Torch-Choice instead dispatches coefficient modules by variable group and accumulates them in a newly allocated dense utility tensor at every forward pass. This repeated dispatch and allocation adds work to each likelihood evaluation. TorchDCM runtime rises from 0.005 to 0.107 seconds as $|\mathcal{N}|$ increases from 1,000 to 100,000, from 0.024 to 0.337 seconds as $|\mathcal{C}|$ increases from 3 to 20, and from 0.029 to 0.087 seconds as $|\mathcal{K}|$ increases from 4 to 32. The corresponding increases are much larger for most reference estimators, particularly in the sample-size and high-dimensional sweeps. Changing $\rho$ has no monotonic effect on runtime, and the final log likelihoods remain consistent even under near-collinearity. In the joint stress tests, TorchDCM completes the small, medium, and large cases in 0.723, 2.480, and 5.203 seconds. Torch-Choice, SciPy, \texttt{mlogit}, and \texttt{xlogit} also complete all three cases but require substantially more time. Biogeme and \texttt{gmnl} time out from the medium case onward, and Apollo times out in the large case.

\begin{table}[htb]
\caption{Synthetic controlled NL runtime comparison (seconds).}
\label{tab:generated_nl}
\centering
\scriptsize
\renewcommand{\arraystretch}{0.96}
\setlength{\tabcolsep}{2.4pt}
\begin{tabular*}{\linewidth}{@{\extracolsep{\fill}}ll*{8}{r}l@{}}
\toprule
Sweep & Case & $|\mathcal{N}|$ & $|\mathcal{C}|$ & $|\mathcal{K}|$ & $\rho$ & TorchDCM & Torch-Choice & Biogeme & Apollo & Consistent? \\
\midrule
Sample size & $|\mathcal{N}|=1{,}000$ & 1000 & 4 & 6 & 0.30 & 0.046 & 0.059 & 24.958 & 0.519 & Yes \\
Sample size & $|\mathcal{N}|=10{,}000$ & 10000 & 4 & 6 & 0.30 & 0.070 & 0.143 & 25.894 & 2.052 & Yes \\
Sample size & $|\mathcal{N}|=100{,}000$ & 100000 & 4 & 6 & 0.30 & 0.547 & 1.256 & 37.292 & 29.578 & Yes \\
Alternatives & $|\mathcal{C}|=3$ & 20000 & 3 & 6 & 0.30 & 0.103 & 0.143 & 9.261 & 2.678 & Yes \\
Alternatives & $|\mathcal{C}|=10$ & 20000 & 10 & 6 & 0.30 & 0.288 & 0.577 & Timeout & 14.117 & Yes \\
Alternatives & $|\mathcal{C}|=20$ & 20000 & 20 & 6 & 0.30 & 0.827 & 3.791 & Timeout & 62.563 & Yes \\
Variables & $|\mathcal{K}|=4$ & 20000 & 5 & 4 & 0.30 & 0.127 & 0.271 & 53.393 & 4.020 & Yes \\
Variables & $|\mathcal{K}|=12$ & 20000 & 5 & 12 & 0.30 & 0.172 & 0.368 & 49.493 & 9.711 & Yes \\
Variables & $|\mathcal{K}|=32$ & 20000 & 5 & 32 & 0.30 & 0.278 & 0.628 & 194.646 & 46.946 & Yes \\
Correlation & $\rho=0.00$ & 20000 & 5 & 12 & 0.00 & 0.179 & 0.382 & 56.258 & 9.627 & Yes \\
Correlation & $\rho=0.50$ & 20000 & 5 & 12 & 0.50 & 0.168 & 0.360 & 48.845 & 11.312 & Yes \\
Correlation & $\rho=0.98$ & 20000 & 5 & 12 & 0.98 & 0.168 & 0.362 & 47.524 & 10.094 & Yes \\
Stress & Small & 30000 & 12 & 8 & 0.50 & 0.552 & 1.191 & Timeout & 37.064 & Yes \\
Stress & Medium & 40000 & 16 & 10 & 0.50 & 1.338 & 6.102 & Timeout & 99.092 & Yes \\
Stress & Large & 50000 & 20 & 12 & 0.50 & 2.781 & 13.058 & Timeout & 227.469 & Yes \\
\bottomrule
\end{tabular*}
\end{table}

\begin{table}[htb]
\caption{Synthetic controlled MixL runtime comparison (seconds).}
\label{tab:generated_mixl}
\centering
\scriptsize
\renewcommand{\arraystretch}{0.96}
\setlength{\tabcolsep}{2.8pt}
\begin{tabular*}{\linewidth}{@{\extracolsep{\fill}}llrrrrrrrrl@{}}
\toprule
Sweep & Case & $|\mathcal{N}|$ & $|\mathcal{C}|$ & $|\mathcal{K}|$ & $\rho$ & $|\mathcal{K}^{\mathrm{R}}|$ & TorchDCM & Biogeme & Apollo & Consistent? \\
\midrule
Sample size & $|\mathcal{N}|=1{,}000$ & 1000 & 4 & 6 & 0.30 & 3 & 0.147 & 58.751 & 2.755 & Yes \\
Sample size & $|\mathcal{N}|=10{,}000$ & 10000 & 4 & 6 & 0.30 & 3 & 0.903 & 67.465 & 37.059 & Yes \\
Sample size & $|\mathcal{N}|=100{,}000$ & 100000 & 4 & 6 & 0.30 & 3 & 14.797 & 172.664 & 257.927 & Yes \\
Alternatives & $|\mathcal{C}|=3$ & 20000 & 3 & 6 & 0.30 & 3 & 1.800 & 59.629 & 38.472 & Yes \\
Alternatives & $|\mathcal{C}|=10$ & 20000 & 10 & 6 & 0.30 & 3 & 7.378 & Timeout & 231.524 & Yes \\
Alternatives & $|\mathcal{C}|=20$ & 20000 & 20 & 6 & 0.30 & 3 & 18.295 & Timeout & Timeout & N.A. \\
Variables & $|\mathcal{K}|=4$ & 20000 & 5 & 4 & 0.30 & 2 & 1.566 & 66.992 & 51.871 & Yes \\
Variables & $|\mathcal{K}|=12$ & 20000 & 5 & 12 & 0.30 & 6 & 3.273 & Timeout & 285.949 & Yes \\
Variables & $|\mathcal{K}|=32$ & 20000 & 5 & 32 & 0.30 & 16 & 8.846 & Timeout & Timeout & N.A. \\
Correlation & $\rho=0.00$ & 20000 & 5 & 12 & 0.00 & 6 & 3.310 & Timeout & 253.278 & Yes \\
Correlation & $\rho=0.50$ & 20000 & 5 & 12 & 0.50 & 6 & 4.764 & Timeout & 249.469 & Yes \\
Correlation & $\rho=0.98$ & 20000 & 5 & 12 & 0.98 & 6 & 5.048 & Timeout & 295.729 & Yes \\
Stress & Small & 20000 & 12 & 8 & 0.50 & 4 & 16.210 & Timeout & Timeout & N.A. \\
Stress & Medium & 30000 & 16 & 10 & 0.50 & 5 & 29.138 & Timeout & Timeout & N.A. \\
Stress & Large & 40000 & 20 & 12 & 0.50 & 6 & 74.576 & Timeout & Timeout & N.A. \\
\bottomrule
\end{tabular*}
\par\vspace{5pt}
\parbox{\linewidth}{\scriptsize\textit{Notes.} N.A.: fewer than two comparable estimates remain.}
\end{table}

The NL cases use two nests. Table~\ref{tab:generated_nl} shows the same overall pattern for NL: TorchDCM is the fastest estimator in every case and completes all cases in 2.781 seconds or less. Torch-Choice ranks second in all 15 cases, with runtimes from 0.059 to 13.058 seconds. Runtime increases most clearly with sample size, choice-set width, and the number of coefficients, whereas changing \(\rho\) has little effect. Apollo completes every case but rises from 37.064 to 227.469 seconds across the three stress levels. Biogeme completes the smaller and lower-dimensional cases but times out for \(|\mathcal{C}|=10\) and 20 and for all stress cases, where the larger number of alternative rows and nested probability calculations exceed the time limit. The final-likelihood criterion is satisfied by every completed comparison reported here.

MixL is more computationally demanding because each likelihood evaluation integrates choice probabilities over simulation draws. Table~\ref{tab:generated_mixl} shows that TorchDCM remains the fastest estimator in every completed comparison, although its runtime rises more sharply than in MNL and NL. Increasing \(|\mathcal{N}|\) from 1,000 to 100,000 raises runtime from 0.147 to 14.797 seconds, while wider choice sets and more random coefficients produce similar increases. TorchDCM completes the three stress cases in 16.210, 29.138, and 74.576 seconds. Both external estimators time out in all three. Biogeme also times out in most cases with wider choice sets or more random coefficients, whereas Apollo completes several of these cases but approaches the 300-second limit. The timeouts reflect the combined computational burden of alternative rows, simulation draws, and random-coefficient dimensions. All solvers that complete within the time limit are included in the final log-likelihood comparison. Every row with at least two such solvers satisfies the final log-likelihood criterion. The \(|\mathcal{C}|=20\) and \(|\mathcal{K}|=32\) cases, together with the three stress cases, are reported as N.A. because \torchdcm is the only solver that completes within the time limit.

\subsection{CPU--GPU comparison on synthetic data}
\label{sec:device_results}

We isolate device execution using identical generated data, specifications, and initial values for MNL, NL, and MixL on CPU and CUDA. MixL also uses identical simulation draws. Reported runtimes are medians of three estimation runs.

\begin{table}[htb]
\caption{TorchDCM CPU--GPU runtime comparison (seconds).}
\label{tab:torch_device_stress}
\centering
\scriptsize
\renewcommand{\arraystretch}{0.96}
\setlength{\tabcolsep}{3pt}
\begin{tabular*}{\linewidth}{@{\extracolsep{\fill}}lrrrrrrrrl@{}}
\toprule
Model & $|\mathcal{N}|$ & $|\mathcal{C}|$ & $|\mathcal{K}|$ & $|\mathcal{K}^{\mathrm{R}}|$ & $R$ & CPU & GPU & Speedup & Consistent? \\
\midrule
MNL & 250000 & 12 & 12 & -- & -- & 2.994 & 0.210 & 14.3$\times$ & Yes \\
MNL & 500000 & 12 & 12 & -- & -- & 8.402 & 0.383 & 22.0$\times$ & Yes \\
MNL & 1000000 & 12 & 12 & -- & -- & 17.247 & 0.752 & 22.9$\times$ & Yes \\
NL & 100000 & 12 & 12 & -- & -- & 2.015 & 0.168 & 12.0$\times$ & Yes \\
NL & 250000 & 12 & 12 & -- & -- & 6.083 & 0.326 & 18.7$\times$ & Yes \\
NL & 500000 & 12 & 12 & -- & -- & 17.942 & 0.610 & 29.4$\times$ & Yes \\
MixL & 5000 & 12 & 12 & 8 & 256 & 23.910 & 0.545 & 43.9$\times$ & Yes \\
MixL & 10000 & 12 & 12 & 8 & 256 & 46.876 & 0.833 & 56.3$\times$ & Yes \\
MixL & 25000 & 12 & 12 & 8 & 256 & 129.483 & 1.823 & 71.0$\times$ & Yes \\
\bottomrule
\end{tabular*}
\end{table}

Table~\ref{tab:torch_device_stress} shows that GPU execution is faster in all nine cases, with identical CPU and GPU estimates. Speedup increases with problem size. It rises from 14.3$\times$ to 22.9$\times$ for MNL and from 12.0$\times$ to 29.4$\times$ for NL. MixL benefits most, with speedup increasing from 43.9$\times$ at 5,000 observations to 71.0$\times$ at 25,000 observations, because its simulated likelihood provides substantially more parallel work per observation. The largest MNL, NL, and MixL cases use 4.0, 2.2, and 3.2 GB of GPU memory, respectively. No device run fails or violates the consistency tolerances.

\subsection{Actual-data results}

The final experiments evaluate MNL, NL, and MixL on the public datasets summarized in Table~\ref{tab:actual_data}. They assess both numerical agreement with aligned reference estimators and full-estimation runtime under empirical data layouts. In Tables~\ref{tab:nested_real} and \ref{tab:mixed_real}, \(\lvert\mathcal{K}\rvert\) counts model-specific utility or mean coefficients. NL dissimilarity parameters and MixL random-coefficient scales are reported separately and are not included in this count.
\begin{table}[htb]
\caption{Real-data MNL runtime comparison (seconds).}
\label{tab:mnl_real}
\centering
\scriptsize
\renewcommand{\arraystretch}{0.96}
\setlength{\tabcolsep}{1.6pt}
\begin{tabular*}{\linewidth}{@{\extracolsep{\fill}}l*{11}{r}l@{}}
\toprule
Data & $|\mathcal{N}|$ & $|\mathcal{C}|$ & $|\mathcal{K}|$ & TorchDCM & Torch-Choice & SciPy & Biogeme & Apollo & \texttt{mlogit} & \texttt{gmnl} & \texttt{xlogit} & Consistent? \\
\midrule
Swissmetro & 10719 & 3 & 4 & 0.010 & 0.014 & 2.366 & 1.346 & 0.725 & 0.123 & Fail & 0.026 & Yes \\
NHTS 2022 & 27375 & 5 & 16 & 0.095 & 0.385 & 38.723 & 2.749 & 7.617 & 1.388 & 18.450 & 0.713 & Yes \\
Airline itinerary & 3609 & 3 & 5 & 0.007 & 0.010 & 3.150 & 1.573 & 0.330 & 0.052 & 0.107 & 0.011 & Yes \\
Parking Spain & 1576 & 3 & 5 & 0.007 & 0.009 & 1.930 & 1.503 & 0.286 & 0.031 & 0.056 & 0.006 & Yes \\
Telephone service & 434 & 5 & 5 & 0.005 & 0.006 & 0.242 & 1.652 & 0.253 & 0.021 & Fail & 0.004 & Yes \\
LPMC London & 81086 & 4 & 5 & 0.088 & 0.201 & 56.996 & 1.575 & 12.044 & 2.020 & 5.644 & 0.334 & Yes \\
Catsup & 2798 & 4 & 3 & 0.004 & 0.009 & 0.376 & 1.592 & 0.271 & 0.049 & 0.049 & 0.008 & Yes \\
Cracker & 3292 & 4 & 3 & 0.009 & 0.017 & 0.545 & 1.570 & 0.275 & 0.055 & 0.056 & 0.009 & Yes \\
Electricity & 4308 & 4 & 6 & 0.012 & 0.021 & 2.861 & 2.031 & 0.406 & 0.070 & 0.242 & 0.013 & Yes \\
HC & 250 & 7 & 2 & 0.004 & 0.005 & 0.048 & 2.157 & 0.215 & 0.014 & 0.017 & 0.002 & Yes \\
Heating & 900 & 5 & 2 & 0.003 & 0.004 & 0.141 & 1.674 & 0.220 & 0.023 & 0.022 & 0.003 & Yes \\
Mode & 453 & 4 & 2 & 0.003 & 0.003 & 0.201 & 1.615 & 0.209 & 0.013 & 0.015 & 0.002 & Yes \\
NOx & 632 & 15 & 3 & 0.005 & 0.008 & 0.236 & 5.146 & 0.286 & 0.035 & Fail & 0.004 & Yes \\
RiskyTransport & 1793 & 4 & 7 & 0.025 & 0.029 & 1.139 & 2.325 & 0.346 & 0.040 & Fail & Fail & Yes \\
Train & 2929 & 2 & 4 & 0.016 & 0.019 & 1.766 & 1.359 & 0.286 & 0.031 & 0.047 & 0.005 & Yes \\
Fishing & 1182 & 4 & 5 & 0.011 & 0.015 & 0.815 & 1.543 & 0.274 & 0.941 & 0.549 & 0.006 & Yes \\
ModeCanada & 4324 & 4 & 3 & 0.009 & 0.014 & 1.967 & 1.463 & 0.332 & 0.521 & Fail & Fail & Yes \\
\bottomrule
\end{tabular*}
\end{table}

Table~\ref{tab:mnl_real} shows that TorchDCM and Torch-Choice complete all 17 real-data MNL cases. Based on the unrounded measurements, TorchDCM is faster than Torch-Choice throughout, reduces median runtime by 30.6\%, and is the fastest estimator overall in nine cases, including NHTS, LPMC, and both ragged datasets. \texttt{xlogit} is fastest in the other eight, primarily small balanced cases. SciPy, Biogeme, Apollo, \texttt{mlogit}, and \texttt{gmnl} generally require more time, with the largest differences occurring when the sample or coefficient dimension is large. All successfully completed estimators satisfy the numerical tolerances. The missing \texttt{gmnl} results arise from non-conformable-array errors on the affected datasets. \texttt{xlogit} does not complete RiskyTransport and ModeCanada because these datasets have ragged choice sets, whereas its MNL interface requires a consistent set of alternatives in long format.

Table~\ref{tab:nested_real} shows that TorchDCM has the lowest runtime in all 12 real-data NL cases, ranging from 0.026 to 0.662 seconds, followed by Torch-Choice at 0.032--1.519 seconds. The NL battery contains fewer datasets because it includes only cases for which a nontrivial nesting structure can be specified. In particular, a nontrivial nesting structure requires at least three alternatives.

\begin{table}[!htb]
\caption{Real-data NL runtime comparison (seconds).}
\label{tab:nested_real}
\centering
\scriptsize
\renewcommand{\arraystretch}{0.96}
\setlength{\tabcolsep}{2.6pt}
\begin{tabular*}{\linewidth}{@{\extracolsep{\fill}}l*{7}{r}l@{}}
\toprule
Data & $|\mathcal{N}|$ & $|\mathcal{C}|$ & $|\mathcal{K}|$ & TorchDCM & Torch-Choice & Biogeme & Apollo & Consistent? \\
\midrule
Swissmetro & 10719 & 3 & 4 & 0.054 & 0.074 & 5.645 & 1.078 & Yes \\
LPMC London & 81086 & 4 & 5 & 0.662 & 1.519 & 23.963 & 13.544 & Yes \\
NHTS 2022 & 27375 & 5 & 16 & 0.457 & 1.004 & 104.413 & Fail & Yes \\
Parking Spain & 1576 & 3 & 5 & 0.026 & 0.032 & 6.555 & 0.391 & Yes \\
Airline itinerary & 3609 & 3 & 5 & 0.059 & 0.076 & 6.523 & 0.389 & Yes \\
Catsup & 2798 & 4 & 3 & 0.043 & 0.064 & 10.606 & 0.345 & Yes \\
Cracker & 3292 & 4 & 3 & 0.031 & 0.046 & 10.573 & 0.378 & Yes \\
Electricity & 4308 & 4 & 6 & 0.082 & 0.144 & 20.628 & Fail & Yes \\
Fishing & 1182 & 4 & 2 & 0.058 & 0.084 & 9.708 & 0.324 & Yes \\
HC & 250 & 7 & 2 & 0.063 & 0.079 & 34.202 & 0.322 & Yes \\
Heating & 900 & 5 & 2 & 0.060 & 0.080 & 10.125 & 0.306 & Yes \\
Mode & 453 & 4 & 2 & 0.027 & 0.035 & 9.796 & 0.295 & Yes \\
\bottomrule
\end{tabular*}
\end{table}

Biogeme requires 5.645--104.413 seconds, while Apollo requires 0.295--13.544 seconds when it completes successfully. The largest runtimes occur for NHTS and LPMC, which have larger samples or coefficient dimensions. HC also takes longer for Biogeme despite its small sample, illustrating that choice-set structure and optimization difficulty also affect runtime. All completed NL estimates satisfy the normalized final-log-likelihood tolerance. Apollo is marked as Fail for NHTS because of false convergence and for Electricity because of singular convergence. Both runs are therefore excluded.

For the real-data MixL comparison, the common mean-coefficient starts are the estimates from the aligned MNL specification. Each random-coefficient scale starts at the reciprocal root mean square (RMS), \(1/\operatorname{RMS}(\boldsymbol{X}^{\mathrm{R}}_k)\), so variables measured in different units imply comparable initial utility variation. These natural-scale starting values are supplied to all three estimators. We exclude the preliminary MNL fit from the reported MixL runtimes. Each MixL optimizer may run for up to 300 iterations in these experiments.

Table~\ref{tab:mixed_real} compares the three estimators on 16 MixL cases. TorchDCM is faster than Biogeme in all 16 cases and faster than Apollo in 13 of the 15 cases that Apollo completes successfully. Thirteen TorchDCM runs finish in less than one second. The larger runtimes for LPMC, ModeCanada, and RiskyTransport reflect their sample sizes or more difficult optimization landscapes. Biogeme runtimes range from 24.729 to 282.477 seconds. Apollo runtimes range from 0.366 to 10.002 seconds, while its LPMC run fails after singular convergence. All rows satisfy the final-likelihood criterion. The Apollo solutions for HC and NOx are marked and excluded because Apollo reports relative-function convergence, with final log likelihoods falling 8.48 and 27.07 below the best values in their respective rows.

\begin{table}[!htb]
\caption{Real-data MixL runtime comparison (seconds).}
\label{tab:mixed_real}
\centering
\scriptsize
\renewcommand{\arraystretch}{0.96}
\setlength{\tabcolsep}{2.5pt}
\begin{tabular*}{\linewidth}{@{\extracolsep{\fill}}lrrrrrrrl@{}}
\toprule
Data & $|\mathcal{N}|$ & $|\mathcal{C}|$ & $|\mathcal{K}|$ & $|\mathcal{K}^{\mathrm{R}}|$ & TorchDCM & Biogeme & Apollo & Consistent? \\
\midrule
Swissmetro & 10719 & 3 & 4 & 2 & 0.411 & 24.729 & 8.587 & Yes \\
Airline itinerary & 3609 & 3 & 5 & 3 & 0.183 & 33.840 & 3.954 & Yes \\
Parking Spain & 1576 & 3 & 5 & 3 & 0.102 & 33.537 & 1.874 & Yes \\
Telephone & 434 & 5 & 5 & 2 & 0.035 & 28.633 & 0.780 & Yes \\
LPMC London & 81086 & 4 & 5 & 2 & 17.478 & 73.769 & Fail & Yes \\
Catsup & 2798 & 4 & 3 & 3 & 0.231 & 43.311 & 2.465 & Yes \\
Cracker & 3292 & 4 & 3 & 3 & 0.298 & 43.987 & 3.194 & Yes \\
Electricity & 4308 & 4 & 6 & 4 & 0.565 & 93.019 & 10.002 & Yes \\
Fishing & 1182 & 4 & 2 & 2 & 0.071 & 28.844 & 0.676 & Yes \\
HC & 250 & 7 & 2 & 2 & 0.045 & 103.683 & 0.366$^{\dagger}$ & Yes \\
Heating & 900 & 5 & 2 & 2 & 0.109 & 77.211 & 0.715 & Yes \\
Mode & 453 & 4 & 2 & 2 & 0.033 & 28.581 & 0.367 & Yes \\
ModeCanada & 4324 & 4 & 4 & 4 & 65.396 & 170.166 & 5.444 & Yes \\
NOx & 632 & 15 & 3 & 3 & 0.151 & 282.477 & 2.137$^{\dagger}$ & Yes \\
RiskyTransport & 1793 & 4 & 7 & 4 & 17.061 & 84.348 & 7.132 & Yes \\
Train & 2929 & 2 & 4 & 4 & 0.474 & 40.851 & 1.276 & Yes \\
\bottomrule
\end{tabular*}
\par\vspace{5pt}
\parbox{\linewidth}{\scriptsize\textit{Notes.} $^{\dagger}$: final log likelihood is below the row best by more than $\tau_{\ell}$ and is excluded from consistency.}
\end{table}

\section{Conclusion}
\label{sec:conclusion}

This paper introduces \torchdcm, a unified PyTorch-native package for discrete choice modeling that connects tensor-native computation with data structures, model specifications, parameter transformations, estimation, inference, prediction, simulation, and reporting. Across the aligned synthetic and real-data experiments, \torchdcm combines comparable final log likelihoods with substantially lower runtimes than Biogeme and Apollo. It completes every synthetic stress case, and its unified CPU/CUDA execution path provides further gains for large and simulation-intensive settings. These results show that tensor-native execution can improve computational scaling while retaining a complete econometric workflow.

The findings apply to the tested models, data, initialization rules, and hardware. Their broader value lies in reducing the implementation and verification cost of scalable choice-model research. The unified package interfaces allow new differentiable likelihood components to reuse data handling, constrained optimization, covariance calculation, prediction, and reporting. The separate public validation pipeline aligns specifications, runtime scope, and numerical diagnostics across software, making new estimators easier to reproduce, audit, and compare. Together, these resources provide an extensible foundation for developing and validating large-scale DCMs.

\bibliographystyle{informs2014}
\bibliography{reference_torchdcm}

\renewcommand{\scriptsize}{\fontsize{7.25}{8.25}\selectfont}
\setlist[itemize]{label=\textbullet,nosep,topsep=4pt}
\renewcommand{\arraystretch}{0.68}
\setlength{\textfloatsep}{5pt plus 2pt minus 2pt}
\setlength{\floatsep}{3pt plus 2pt minus 2pt}
\setlength{\intextsep}{3pt plus 2pt minus 2pt}
\setlength{\abovedisplayskip}{3.5pt plus 2pt minus 2pt}
\setlength{\belowdisplayskip}{3.5pt plus 2pt minus 2pt}
\setlength{\abovedisplayshortskip}{3pt plus 2pt minus 2pt}
\setlength{\belowdisplayshortskip}{3.5pt plus 2pt minus 2pt}
\AtBeginEnvironment{table}{\TableSpaced}
\lstdefinestyle{torchdcmPython}{
    linewidth=\linewidth,
    language=Python,
    basicstyle=\scriptsize\ttfamily,
    keywordstyle=\color{codekeyword}\bfseries,
    commentstyle=\color{codecomment}\itshape,
    stringstyle=\color{codestring},
    morekeywords=[2]{torch,report,to_text,save_report},
    keywordstyle=[2]\color{codeapi}\bfseries,
    morekeywords=[3]{True,False,None},
    keywordstyle=[3]\color{codeconstant}\bfseries,
    backgroundcolor=\color{codebg},
    rulecolor=\color{codeframe},
    columns=flexible,
    keepspaces=true,
    showstringspaces=false,
    breaklines=true,
    breakatwhitespace=true,
    tabsize=2,
    numbers=none,
    captionpos=t,
    aboveskip=0pt,
    belowskip=0pt,
    lineskip=-1pt,
    frame=single,
    framesep=5pt,
    xleftmargin=0pt,
    xrightmargin=0pt,
}
\lstset{style=torchdcmPython}

\RUNAUTHOR{Mo et al.}
\RUNTITLE{TorchDCM: A Unified PyTorch-Native Package}

\ECRUNAUTHOR{Mo et al.}
\ECAUpunct{}
\ECSwitch
\providecommand{\theHsection}{}
\providecommand{\theHsubsection}{}
\providecommand{\theHsubsubsection}{}
\providecommand{\theHequation}{}
\providecommand{\theHfigure}{}
\providecommand{\theHtable}{}
\renewcommand{\theHsection}{EC.\arabic{section}}
\renewcommand{\theHsubsection}{EC.\arabic{section}.\arabic{subsection}}
\renewcommand{\theHsubsubsection}{EC.\arabic{section}.\arabic{subsection}.\arabic{subsubsection}}
\renewcommand{\theHequation}{EC.\arabic{equation}}
\renewcommand{\theHfigure}{EC.\arabic{figure}}
\renewcommand{\theHtable}{EC.\arabic{table}}
\ECHead{Electronic Companion}
\setcounter{lstlisting}{0}
\renewcommand{\thelstlisting}{EC.\arabic{lstlisting}}
\providecommand{\theHlstlisting}{}
\renewcommand{\theHlstlisting}{EC.\arabic{lstlisting}}

\looseness=-1
The EC extends the main paper in three areas: package design, implementation,
and benchmark evidence.
Section~\ref{ec:estimation_report} documents the structured estimation report
and its export formats. Section~\ref{ec:ordered_response} establishes
full-output agreement for ordered logit and ordered probit through likelihood,
parameter, and prediction comparisons. Section~\ref{ec:advanced_likelihood}
then reports full-estimation validation for latent-class, hybrid-choice, and
panel models.

\section{Structured Estimation Report and Example}
\label{ec:estimation_report}

\looseness=-1
\torchdcm uses a unified report to support a consistent audit of a fitted model,
from specification and optimization through inference and post-estimation. A \texttt{ChoiceResults}
object retains the numerical results, fitted model, and data. Calling
\texttt{report} converts this information into an \texttt{EstimationReport}
whose schema defines the following report contract:

\begin{itemize}
    \item model, package version, device, dtype, optimizer, and random-seed
    fields document the computational setting for reproducible export;
    \item individual, choice-observation, alternative-row, availability,
    weighting, panel, and balanced-versus-ragged fields support data and
    specification checks;
    \item nests, random-coefficient distributions, draw counts, and panel
    integration settings verify how each model is constructed;
    \item closure evaluations, gradient norms, information-matrix diagnostics,
    and component runtimes support convergence diagnosis;
    \item likelihood statistics, covariance type, alternative shares, grouped
    estimates, and numerical warnings support inference and post-estimation.
\end{itemize}

\looseness=-1
The parameter table reports estimates, standard errors, reference values, asymptotic \(z\)-statistics, \(p\)-values, and confidence intervals. Fixed parameters remain visible and are labeled rather than assigned artificial standard errors. Multinomial logit (MNL) utility coefficients use the reference value zero. Nested logit (NL) dissimilarity parameters use one because this is the value at which the nest reduces to the corresponding MNL structure. Mixed logit (MixL) output separates coefficient means, scales, and Cholesky parameters.

Listing~\ref{ec:lst:report_export} shows the report workflow. HTML and text
support direct inspection. JSON preserves the complete schema, CSV files store
tabular outputs, and the \LaTeX{} fragment can be incorporated into an
application report.

\begin{lstlisting}[caption={Generating and exporting a single-model estimation report.},label={ec:lst:report_export}]
result = model.fit(data, cov_type="cluster", groups="person_id")
report = result.report(
    cov_type="cluster",
    confidence_level=0.95,
)
print(report.to_text())
result.save_report(
    "outputs/swissmetro_mnl",
    formats=["html", "json", "csv", "latex", "text"],
)
\end{lstlisting}

\begin{figure}[htb]
\centering
\includegraphics[width=0.56\linewidth]{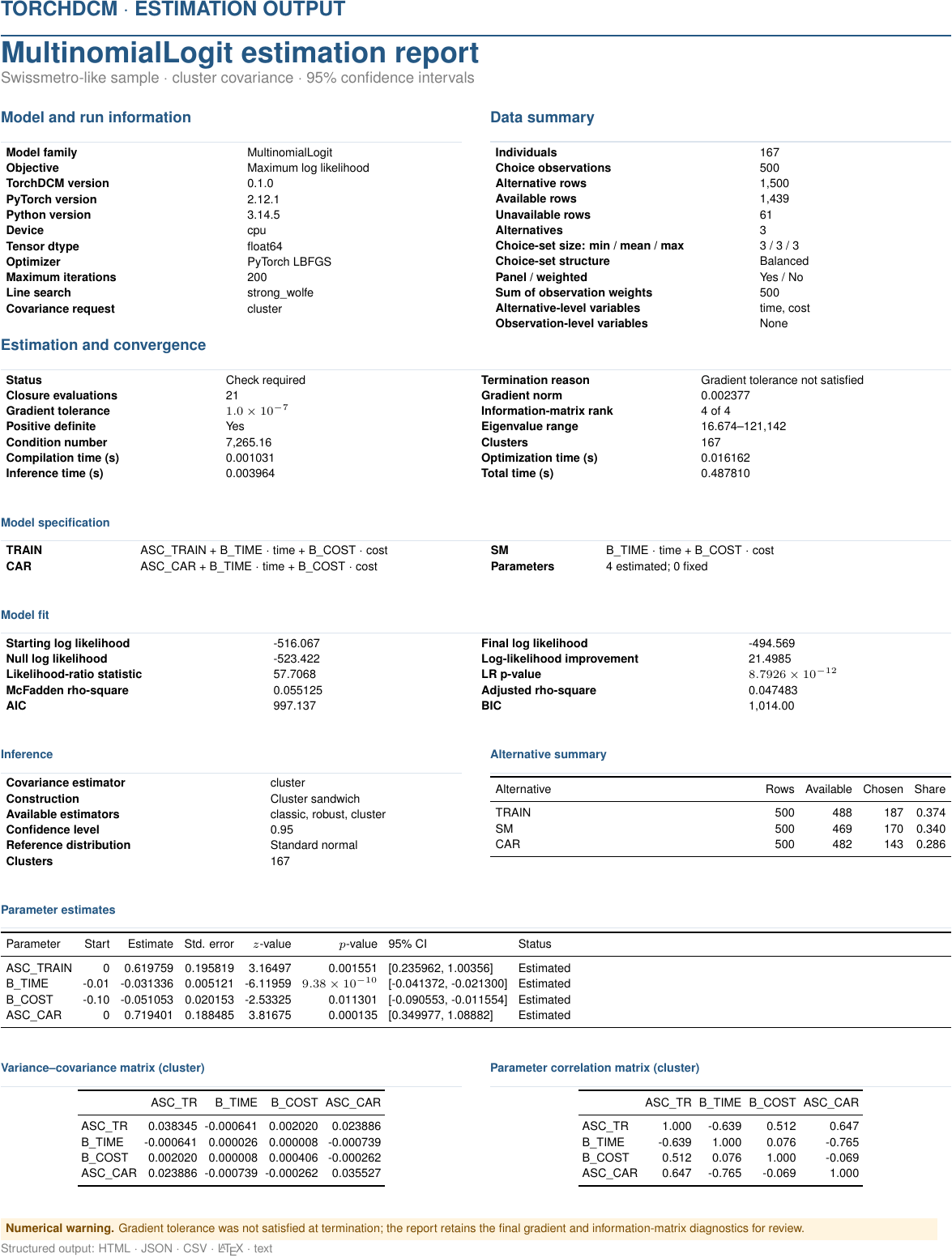}
\caption{An example of the TorchDCM estimation report.}
\label{ec:fig:report_excerpt}
\end{figure}

The exported report is available as HTML (\texttt{report.html}), JSON
(\texttt{result.json}), text (\texttt{summary.txt}), and a \LaTeX{} fragment
(\texttt{report.tex}). Four CSV files store the parameter table
(\texttt{parameters.csv}), alternative summary (\texttt{alternatives.csv}),
covariance matrix (\texttt{covariance.csv}), and correlation matrix
(\texttt{correlation.csv}). Figure~\ref{ec:fig:report_excerpt} shows a one-page
vector rendering of the core report for the 500-observation Swissmetro-like MNL
example. It includes model and data summaries, convergence diagnostics, fit
statistics, inference settings, alternative shares, parameter estimates,
covariance and correlation matrices, and numerical warnings. The exported HTML
and machine-readable files retain the complete report.

\section{Ordered Logit and Ordered Probit Validation}
\label{ec:ordered_response}

\looseness=-1
We validate the ordered-logit and ordered-probit estimators against both Biogeme
and Apollo. The experiments cover full maximum-likelihood estimation,
constrained thresholds, covariance calculation, and prediction for controlled
synthetic cases and all ordered-response indicators in the aligned Optima
benchmark.

\subsection{Model and validation protocol}

Let \(\mathcal{N}\) be the set of observations, \(\mathcal{K}\) the set of covariates, and \(\mathcal{C}=\{1,\ldots,\lvert\mathcal{C}\rvert\}\) the ordered response levels. Both models introduce the latent response
\begin{align}
    y_n^*
    &=\boldsymbol{x}_n^{\mathsf T}\boldsymbol{\beta}+\varepsilon_n,
    \qquad n\in\mathcal{N},
    \label{ec:eq:ordered_latent}
\end{align}
and map it to the observed response through increasing thresholds,
\begin{align}
    y_n=c
    &\quad\Longleftrightarrow\quad
    \tau_{c-1}<y_n^*\leq\tau_c,
    &c&\in\mathcal{C},
    \label{ec:eq:ordered_mapping}
\end{align}
where \(\tau_0=-\infty\), \(\tau_{\lvert\mathcal{C}\rvert}=+\infty\), and \(\tau_1<\cdots<\tau_{\lvert\mathcal{C}\rvert-1}\). The response probability is
\begin{align}
    P_{n,c}
    &=F\!\left(\tau_c-
      \boldsymbol{x}_n^{\mathsf T}\boldsymbol{\beta}\right)
      -F\!\left(\tau_{c-1}-
      \boldsymbol{x}_n^{\mathsf T}\boldsymbol{\beta}\right),
    \qquad c\in\mathcal{C}.
    \label{ec:eq:ordered_probability}
\end{align}
\looseness=-1
Eqs.~\eqref{ec:eq:ordered_latent}--\eqref{ec:eq:ordered_probability} define both
models, with \(F=\Lambda\), the standard-logistic cumulative distribution
function, for ordered logit and \(F=\Phi\), the standard-normal cumulative
distribution function, for ordered probit. \torchdcm evaluates both through the
same vectorized ordered-response kernel. It uses an unrestricted first threshold
and constructs subsequent thresholds by cumulatively adding exponentiated
internal gap parameters.

\looseness=-1
All experiments use the single-CPU setup defined in the article. Runtime
includes estimation and covariance construction but excludes data generation
or loading, process startup, and file input/output. All three estimators receive
identical data and specifications. The synthetic cases also use common
natural-scale starting values. In the Optima cases, threshold starting values
are shared, whereas coefficient starting values are set by the corresponding
package wrappers.

\looseness=-1
Ordered-response validation assesses \emph{full-output agreement} across the
final log likelihood, natural-scale parameters, and predicted probabilities.
Let \(\mathcal{B}\) denote the three estimators. For estimator
\(b\), the vector \(\boldsymbol{P}_b\) stacks \(P_{n,c}\) in observation and
response-level order. We report
\begin{align}
    \Delta_{\ell}
    &=\max_{b,b'\in\mathcal{B}}\lvert\ell_b-\ell_{b'}\rvert,
    &
    \Delta_{\theta}
    &=\max_{b,b'\in\mathcal{B}}
      \lVert\boldsymbol{\theta}_b-\boldsymbol{\theta}_{b'}\rVert_{\infty},
    &
    \Delta_{P}
    &=\max_{b,b'\in\mathcal{B}}
      \lVert\boldsymbol{P}_b-\boldsymbol{P}_{b'}\rVert_{\infty}.
    \label{ec:eq:ordered_agreement}
\end{align}
Full-output agreement requires
\(\Delta_{\ell}\leq\max\{10^{-5},10^{-8}
\max_{b\in\mathcal{B}}\lvert\ell_b\rvert\}\),
\(\Delta_{\theta}\leq10^{-3}\), and \(\Delta_P\leq10^{-4}\).
Thresholds and all other parameters are compared on their natural scales. In
the combined tables, ``Yes'' requires both ordered logit and ordered probit to
pass.

\subsection{Synthetic-data validation}

We apply these criteria to controlled synthetic cases. The generator draws
\(\boldsymbol{x}_n\sim\mathcal{N}(\boldsymbol{0},\boldsymbol{\Sigma}_{\rho})\),
where
\begin{align}
    \boldsymbol{\Sigma}_{\rho}
    &=(1-\rho)\boldsymbol{I}_{\lvert\mathcal{K}\rvert}
      +\rho\boldsymbol{1}_{\lvert\mathcal{K}\rvert}
      \boldsymbol{1}_{\lvert\mathcal{K}\rvert}^{\mathsf T}.
    \label{ec:eq:ordered_covariates}
\end{align}
\looseness=-1
In Eq.~\eqref{ec:eq:ordered_covariates},
\(\boldsymbol{I}_{\lvert\mathcal{K}\rvert}\) is the identity matrix and
\(\boldsymbol{1}_{\lvert\mathcal{K}\rvert}\) is the all-ones vector. Writing
\(a_k=0.4+0.6(k-1)/(\lvert\mathcal{K}\rvert-1)\), the generator sets
\(\beta_k=0.9(-1)^{k-1}a_k/\lVert\boldsymbol{a}\rVert_2\) for
\(k=1,\ldots,\lvert\mathcal{K}\rvert\). It then draws an independent
100,000-observation calibration sample from the same covariate distribution and
sets \(\tau_k\) to the empirical \(k/\lvert\mathcal{C}\rvert\) quantile of its
latent response for \(k=1,\ldots,\lvert\mathcal{C}\rvert-1\). The
data-generating error is standard logistic for ordered logit and standard
normal for ordered probit. The ten rows use seeds
\(20260721,\ldots,20260730\) in the order shown in
Table~\ref{ec:tab:ordered_synthetic}. Estimation begins from zero coefficients
and thresholds implied by the empirical cumulative response shares. The rows
form a case grid over sample size, response levels, covariate dimension, and
equicorrelation; the joint case increases all dimensions.

\begin{table}[htb]
\caption{Synthetic ordered-response runtime comparison (seconds).}
\label{ec:tab:ordered_synthetic}
\centering
\scriptsize
\setlength{\tabcolsep}{1.6pt}
\begin{tabular*}{\linewidth}{@{\extracolsep{\fill}}lrrrrrrrrrrl@{}}
\toprule
& & & & & \multicolumn{3}{c}{Ordered logit} &
\multicolumn{3}{c}{Ordered probit} & \\
\cmidrule(lr){6-8}\cmidrule(lr){9-11}
Case & $\lvert\mathcal{N}\rvert$ & $\lvert\mathcal{C}\rvert$ &
$\lvert\mathcal{K}\rvert$ & $\rho$ & TorchDCM & Biogeme & Apollo &
TorchDCM & Biogeme & Apollo & \makecell{Full-output\\agreement?} \\
\midrule
Sample small & 500 & 6 & 4 & 0.00 & 0.140 & 4.248 & 0.243 & 0.142 & 4.538 & 0.351 & Yes \\
Baseline & 2,000 & 6 & 4 & 0.00 & 0.126 & 5.448 & 0.288 & 0.133 & 5.739 & 0.474 & Yes \\
Sample large & 10,000 & 6 & 4 & 0.00 & 0.134 & 11.774 & 1.073 & 0.143 & 11.986 & 1.556 & Yes \\
Levels low & 2,000 & 3 & 4 & 0.00 & 0.126 & 2.320 & 0.334 & 0.125 & 2.284 & 0.340 & Yes \\
Levels high & 2,000 & 10 & 4 & 0.00 & 0.135 & 13.163 & 0.382 & 0.146 & 14.642 & 0.811 & Yes \\
Variables low & 2,000 & 6 & 2 & 0.00 & 0.127 & 5.246 & 0.262 & 0.131 & 5.514 & 0.382 & Yes \\
Variables high & 2,000 & 6 & 12 & 0.00 & 0.130 & 6.312 & 0.434 & 0.139 & 6.684 & 1.103 & Yes \\
Correlation medium & 2,000 & 6 & 4 & 0.50 & 0.127 & 5.445 & 0.288 & 0.134 & 5.754 & 0.469 & Yes \\
Correlation high & 2,000 & 6 & 4 & 0.90 & 0.129 & 5.461 & 0.291 & 0.129 & 5.711 & 0.466 & Yes \\
Joint large & 20,000 & 10 & 12 & 0.50 & 0.181 & 57.747 & 5.546 & 0.215 & 58.975 & 10.351 & Yes \\
\bottomrule
\end{tabular*}
\vspace{5pt}
\parbox{\linewidth}{\scriptsize\textit{Note.} ``Yes'' denotes full-output
agreement under the maximum pairwise criteria in
Eq.~\eqref{ec:eq:ordered_agreement}.}
\end{table}

\looseness=-1
All 20 model--case combinations satisfy full-output agreement. Across
ordered-logit cases, the largest pairwise likelihood, parameter, and
probability differences are \(1.16\times10^{-7}\),
\(2.11\times10^{-5}\), and \(2.69\times10^{-6}\). The corresponding
ordered-probit maxima are \(5.02\times10^{-8}\),
\(6.93\times10^{-6}\), and \(3.27\times10^{-6}\). All three estimators take
longest in the joint case. Across all cases, TorchDCM remains below 0.22
seconds; in the joint case, Biogeme requires about 58 seconds and Apollo
5.5--10.4 seconds.

\subsection{Actual-data validation}

\looseness=-1
We evaluate 108 model--indicator cases from 54 six-level Likert indicators in
four Optima survey domains. These comprise six
environmental indicators, 27 mobility indicators, seven residential-choice
indicators, and 14 lifestyle indicators. Depending on the indicator, filtering
nonresponses leaves 1,792--1,827 observations. Each model uses the latent index
\begin{align}
    \eta_n
    &=\beta_{\mathrm{male}}\,\mathrm{male}_n
      +\beta_{\mathrm{education}}\,\mathrm{highEducation}_n
      \notag\\
    &\quad+\beta_{\mathrm{GA}}\,\mathrm{haveGA}_n
      +\beta_{\mathrm{income}}\,\mathrm{ScaledIncome}_n.
    \label{ec:eq:optima_latent}
\end{align}
The specification in Eq.~\eqref{ec:eq:optima_latent} also includes five
estimated thresholds and uses the normalized survey weights distributed with
the Biogeme example data.

Table~\ref{ec:tab:ordered_domain_summary} summarizes both ordered models by
domain. Runtime columns give the range across all indicators and both models.
Each difference column reports the largest pairwise value within each of the
four domains.

\begin{table}[htb]
\caption{Ordered-response domain summary.}
\label{ec:tab:ordered_domain_summary}
\centering
\scriptsize
\setlength{\tabcolsep}{2.5pt}
\begin{tabular*}{\linewidth}{@{\extracolsep{\fill}}lrrrrrrr@{}}
\toprule
Domain & Indicators & TorchDCM & Biogeme & Apollo &
\(\max\Delta_{\ell}\) & \(\max\Delta_{\theta}\) &
\(\max\Delta_P\) \\
\midrule
Environmental & 6 & 0.150--0.165 & 5.332--5.660 & 0.330--1.176 &
\(1.13\times10^{-6}\) & \(8.64\times10^{-5}\) & \(1.99\times10^{-5}\) \\
Mobility & 27 & 0.145--0.172 & 5.187--6.289 & 0.317--3.416 &
\(1.40\times10^{-6}\) & \(3.28\times10^{-4}\) & \(3.12\times10^{-5}\) \\
Residential choice & 7 & 0.147--0.167 & 5.309--5.679 & 0.322--0.994 &
\(5.45\times10^{-7}\) & \(1.32\times10^{-4}\) & \(1.15\times10^{-5}\) \\
Lifestyle & 14 & 0.145--0.168 & 5.233--6.062 & 0.324--19.079 &
\(8.53\times10^{-7}\) & \(5.19\times10^{-5}\) & \(1.42\times10^{-5}\) \\
\bottomrule
\end{tabular*}
\vspace{5pt}
\parbox{\linewidth}{\scriptsize\textit{Note.} Runtime ranges are reported in seconds.
Differences use Eq.~\eqref{ec:eq:ordered_agreement}. All 108 cases attain
full-output agreement.}
\end{table}

The following tables report the complete indicator-level results. Their final
column applies the same full-output agreement criteria.

\begin{table}[htb]
\caption{Optima environmental ordered-response runtime (seconds).}
\label{ec:tab:ordered_environmental}
\centering
\scriptsize
\setlength{\tabcolsep}{2.2pt}
\begin{tabular*}{\linewidth}{@{\extracolsep{\fill}}lrrrrrrrl@{}}
\toprule
& & \multicolumn{3}{c}{Ordered logit} &
\multicolumn{3}{c}{Ordered probit} & \\
\cmidrule(lr){3-5}\cmidrule(lr){6-8}
Indicator & $\lvert\mathcal{N}\rvert$ & TorchDCM & Biogeme & Apollo &
TorchDCM & Biogeme & Apollo & \makecell{Full-output\\agreement?} \\
\midrule
Envir01 & 1,822 & 0.159 & 5.352 & 0.330 & 0.165 & 5.660 & 1.003 & Yes \\
Envir02 & 1,818 & 0.150 & 5.332 & 0.381 & 0.158 & 5.645 & 1.006 & Yes \\
Envir03 & 1,807 & 0.155 & 5.349 & 0.348 & 0.155 & 5.640 & 1.176 & Yes \\
Envir04 & 1,823 & 0.154 & 5.364 & 0.332 & 0.158 & 5.660 & 0.951 & Yes \\
Envir05 & 1,821 & 0.159 & 5.365 & 0.336 & 0.165 & 5.639 & 0.960 & Yes \\
Envir06 & 1,822 & 0.155 & 5.410 & 0.384 & 0.162 & 5.646 & 1.082 & Yes \\
\bottomrule
\end{tabular*}
\end{table}

\begin{table}[htb]
\caption{Optima mobility ordered-response runtime, indicators 1--14 (seconds).}
\label{ec:tab:ordered_mobility_1}
\centering
\scriptsize
\setlength{\tabcolsep}{2.2pt}
\begin{tabular*}{\linewidth}{@{\extracolsep{\fill}}lrrrrrrrl@{}}
\toprule
& & \multicolumn{3}{c}{Ordered logit} &
\multicolumn{3}{c}{Ordered probit} & \\
\cmidrule(lr){3-5}\cmidrule(lr){6-8}
Indicator & $\lvert\mathcal{N}\rvert$ & TorchDCM & Biogeme & Apollo &
TorchDCM & Biogeme & Apollo & \makecell{Full-output\\agreement?} \\
\midrule
Mobil01 & 1,810 & 0.156 & 5.350 & 0.333 & 0.160 & 5.607 & 0.871 & Yes \\
Mobil02 & 1,792 & 0.156 & 5.187 & 0.421 & 0.156 & 5.512 & 0.871 & Yes \\
Mobil03 & 1,815 & 0.153 & 5.407 & 0.332 & 0.161 & 5.643 & 1.541 & Yes \\
Mobil04 & 1,819 & 0.152 & 5.377 & 0.328 & 0.155 & 5.695 & 0.908 & Yes \\
Mobil05 & 1,822 & 0.154 & 5.299 & 0.326 & 0.161 & 5.670 & 0.869 & Yes \\
Mobil06 & 1,822 & 0.153 & 5.345 & 0.363 & 0.156 & 5.656 & 0.866 & Yes \\
Mobil07 & 1,817 & 0.161 & 5.383 & 0.336 & 0.169 & 5.639 & 0.885 & Yes \\
Mobil08 & 1,825 & 0.171 & 5.323 & 0.413 & 0.163 & 5.629 & 0.904 & Yes \\
Mobil09 & 1,822 & 0.153 & 5.356 & 0.327 & 0.166 & 5.610 & 0.909 & Yes \\
Mobil10 & 1,801 & 0.158 & 5.350 & 1.257 & 0.159 & 5.664 & 1.067 & Yes \\
Mobil11 & 1,826 & 0.154 & 5.345 & 0.341 & 0.157 & 5.688 & 1.101 & Yes \\
Mobil12 & 1,822 & 0.160 & 5.334 & 0.364 & 0.168 & 5.671 & 1.230 & Yes \\
Mobil13 & 1,823 & 0.159 & 5.413 & 0.341 & 0.162 & 6.289 & 3.416 & Yes \\
Mobil14 & 1,821 & 0.145 & 5.336 & 0.330 & 0.159 & 5.659 & 0.891 & Yes \\
\bottomrule
\end{tabular*}
\end{table}

\begin{table}[htb]
\caption{Optima mobility ordered-response runtime, indicators 15--27 (seconds).}
\label{ec:tab:ordered_mobility_2}
\centering
\scriptsize
\setlength{\tabcolsep}{2.2pt}
\begin{tabular*}{\linewidth}{@{\extracolsep{\fill}}lrrrrrrrl@{}}
\toprule
& & \multicolumn{3}{c}{Ordered logit} &
\multicolumn{3}{c}{Ordered probit} & \\
\cmidrule(lr){3-5}\cmidrule(lr){6-8}
Indicator & $\lvert\mathcal{N}\rvert$ & TorchDCM & Biogeme & Apollo &
TorchDCM & Biogeme & Apollo & \makecell{Full-output\\agreement?} \\
\midrule
Mobil15 & 1,821 & 0.155 & 5.345 & 0.350 & 0.158 & 5.740 & 0.970 & Yes \\
Mobil16 & 1,826 & 0.150 & 5.334 & 0.362 & 0.152 & 5.623 & 0.939 & Yes \\
Mobil17 & 1,818 & 0.158 & 5.343 & 0.331 & 0.162 & 5.607 & 1.017 & Yes \\
Mobil18 & 1,827 & 0.152 & 5.359 & 0.320 & 0.152 & 5.635 & 0.841 & Yes \\
Mobil19 & 1,827 & 0.154 & 5.367 & 0.329 & 0.161 & 5.731 & 0.988 & Yes \\
Mobil20 & 1,816 & 0.172 & 5.210 & 0.337 & 0.158 & 5.561 & 0.902 & Yes \\
Mobil21 & 1,820 & 0.148 & 5.219 & 0.322 & 0.153 & 5.587 & 0.943 & Yes \\
Mobil22 & 1,820 & 0.159 & 5.236 & 0.334 & 0.151 & 5.559 & 0.889 & Yes \\
Mobil23 & 1,812 & 0.151 & 5.265 & 0.348 & 0.156 & 5.571 & 1.031 & Yes \\
Mobil24 & 1,818 & 0.150 & 5.333 & 0.325 & 0.155 & 5.641 & 0.960 & Yes \\
Mobil25 & 1,818 & 0.155 & 5.354 & 0.419 & 0.161 & 5.641 & 0.894 & Yes \\
Mobil26 & 1,821 & 0.153 & 5.342 & 0.344 & 0.160 & 5.679 & 0.877 & Yes \\
Mobil27 & 1,825 & 0.158 & 5.328 & 0.317 & 0.155 & 5.663 & 0.971 & Yes \\
\bottomrule
\end{tabular*}
\end{table}

\begin{table}[htb]
\caption{Optima residential-choice ordered-response runtime (seconds).}
\label{ec:tab:ordered_residential}
\centering
\scriptsize
\setlength{\tabcolsep}{2.2pt}
\begin{tabular*}{\linewidth}{@{\extracolsep{\fill}}lrrrrrrrl@{}}
\toprule
& & \multicolumn{3}{c}{Ordered logit} &
\multicolumn{3}{c}{Ordered probit} & \\
\cmidrule(lr){3-5}\cmidrule(lr){6-8}
Indicator & $\lvert\mathcal{N}\rvert$ & TorchDCM & Biogeme & Apollo &
TorchDCM & Biogeme & Apollo & \makecell{Full-output\\agreement?} \\
\midrule
ResidCh01 & 1,817 & 0.159 & 5.320 & 0.552 & 0.147 & 5.645 & 0.911 & Yes \\
ResidCh02 & 1,817 & 0.156 & 5.326 & 0.337 & 0.162 & 5.625 & 0.869 & Yes \\
ResidCh03 & 1,815 & 0.153 & 5.360 & 0.335 & 0.153 & 5.653 & 0.860 & Yes \\
ResidCh04 & 1,813 & 0.155 & 5.321 & 0.334 & 0.155 & 5.622 & 0.956 & Yes \\
ResidCh05 & 1,813 & 0.153 & 5.343 & 0.357 & 0.167 & 5.679 & 0.994 & Yes \\
ResidCh06 & 1,814 & 0.156 & 5.309 & 0.322 & 0.164 & 5.625 & 0.867 & Yes \\
ResidCh07 & 1,818 & 0.152 & 5.380 & 0.346 & 0.159 & 5.622 & 0.929 & Yes \\
\bottomrule
\end{tabular*}
\end{table}

\begin{table}[htb]
\caption{Optima lifestyle ordered-response runtime (seconds).}
\label{ec:tab:ordered_lifestyle}
\centering
\scriptsize
\setlength{\tabcolsep}{2.2pt}
\begin{tabular*}{\linewidth}{@{\extracolsep{\fill}}lrrrrrrrl@{}}
\toprule
& & \multicolumn{3}{c}{Ordered logit} &
\multicolumn{3}{c}{Ordered probit} & \\
\cmidrule(lr){3-5}\cmidrule(lr){6-8}
Indicator & $\lvert\mathcal{N}\rvert$ & TorchDCM & Biogeme & Apollo &
TorchDCM & Biogeme & Apollo & \makecell{Full-output\\agreement?} \\
\midrule
LifSty01 & 1,820 & 0.157 & 5.233 & 0.468 & 0.163 & 5.539 & 1.093 & Yes \\
LifSty02 & 1,820 & 0.153 & 5.258 & 0.331 & 0.149 & 5.539 & 0.831 & Yes \\
LifSty03 & 1,822 & 0.150 & 5.366 & 0.341 & 0.159 & 5.604 & 0.922 & Yes \\
LifSty04 & 1,810 & 0.161 & 5.357 & 0.331 & 0.168 & 5.603 & 0.914 & Yes \\
LifSty05 & 1,806 & 0.145 & 5.344 & 0.324 & 0.161 & 5.667 & 0.886 & Yes \\
LifSty06 & 1,814 & 0.155 & 5.346 & 0.324 & 0.155 & 5.655 & 0.969 & Yes \\
LifSty07 & 1,812 & 0.153 & 5.283 & 0.327 & 0.162 & 6.062 & 0.878 & Yes \\
LifSty08 & 1,819 & 0.159 & 5.391 & 0.341 & 0.160 & 5.634 & 1.041 & Yes \\
LifSty09 & 1,814 & 0.155 & 5.811 & 0.330 & 0.160 & 5.677 & 1.137 & Yes \\
LifSty10 & 1,814 & 0.155 & 5.319 & 0.325 & 0.155 & 5.623 & 1.203 & Yes \\
LifSty11 & 1,813 & 0.150 & 5.315 & 0.432 & 0.155 & 5.607 & 19.079 & Yes \\
LifSty12 & 1,811 & 0.149 & 5.361 & 1.021 & 0.161 & 5.631 & 0.912 & Yes \\
LifSty13 & 1,819 & 0.155 & 5.408 & 0.390 & 0.159 & 5.638 & 1.054 & Yes \\
LifSty14 & 1,818 & 0.157 & 5.337 & 0.421 & 0.152 & 5.620 & 0.881 & Yes \\
\bottomrule
\end{tabular*}
\end{table}

\FloatBarrier
\looseness=-1
All 108 model--indicator cases attain full-output agreement. The largest
pairwise likelihood,
parameter, and probability differences are \(1.40\times10^{-6}\),
\(3.28\times10^{-4}\), and \(3.12\times10^{-5}\). TorchDCM runtimes are
0.145--0.172 seconds for ordered logit and 0.147--0.169 seconds for ordered
probit. Biogeme requires 5.187--5.811 and 5.512--6.289 seconds, respectively.
Apollo requires 0.317--1.257 and 0.831--19.079 seconds. Apollo's 19.079-second
maximum occurs for LifSty11 ordered probit.

\FloatBarrier
\clearpage
\section{Latent-Class, Hybrid-Choice, and Panel Full-Estimation Validation}
\label{ec:advanced_likelihood}

\looseness=-1
We compare full maximum-likelihood estimation in TorchDCM, Biogeme, and Apollo
on controlled synthetic data and three public datasets. The common protocol is
summarized in Table~\ref{ec:tab:advanced_protocol}. The packages receive
identical data, natural-scale starting values, and explicit antithetic normal
draws for simulated likelihoods.

\begin{table}[htb]
\caption{Advanced-model validation protocol.}
\label{ec:tab:advanced_protocol}
\centering
\scriptsize
\setlength{\tabcolsep}{4pt}
\begin{tabularx}{\linewidth}{@{}p{0.19\linewidth}X@{}}
\toprule
Item & Protocol \\
\midrule
Software & TorchDCM 0.1.2 on PyTorch 2.12.1, Biogeme 3.3.3, and Apollo
0.3.8. \\
Compute & AMD Ryzen 9 9950X3D. Each process and its children use one logical
CPU. Open Multi-Processing (OpenMP), basic linear algebra subprograms (BLAS),
PyTorch intra-operation, and PyTorch inter-operation thread counts equal one.
Apollo uses \texttt{nCores=1}. \\
Data and scale & Controlled data-generating processes and the public Swissmetro, Optima, and
Electricity datasets. Section-specific tables report all sample sizes and draw
counts. We report parameters and starting values on a common natural scale. \\
Limits & At most 150 optimizer iterations for each package and case. A
300-second wall-clock limit applies to the Biogeme and Apollo workers. \\
Runtime boundary & Data generation or loading, conversion to each package,
model construction, process startup, and file input/output are excluded.
Timing begins immediately before \texttt{fit} or the corresponding estimation
call and ends after classic covariance construction. Work performed inside the
estimation call is included; TorchDCM's case-specific compilation and cache
construction occur before the timed call and are excluded. \\
TorchDCM warm-up & One untimed two-iteration latent-class fit with 128
synthetic observations. This warm-up covers compilation, automatic
differentiation, the limited-memory Broyden--Fletcher--Goldfarb--Shanno
(L-BFGS) algorithm, and classic covariance construction and removes their
one-time setup costs. \\
Agreement & Let \(\ell^{\star}\) denote the row-best final log likelihood and
\(\tau_{\ell}=\max\{0.25,10^{-5}|\ell^{\star}|,0.01N_{\mathrm{obs}}\}\), where
\(N_{\mathrm{obs}}\) is the number of choice observations. A completed estimate
more than \(\tau_{\ell}\) below \(\ell^{\star}\) is excluded. ``Yes'' requires
at least two comparable estimates to remain; otherwise, the result is N.A. \\
\bottomrule
\end{tabularx}
\end{table}

In the latent-class and hybrid-choice cases, \(\lvert\mathcal{N}\rvert\)
denotes choice observations. In the panel cases, it denotes individuals, and
\(\lvert\mathcal{O}\rvert\) denotes choice observations.

\looseness=-1
The actual-data cases use deterministic sampling rules. Swissmetro first
removes rows with \texttt{CHOICE}=0 and then takes the requested number of
rows from the remaining data. Optima retains observations with
\(\texttt{Choice}\in\{1,2\}\), positive \texttt{Envir01} and
\texttt{Envir02}, and finite \texttt{TimePT\_scaled},
\texttt{TimeCar\_scaled}, and \texttt{ScaledIncome}. It then shuffles all 1,298
retained observations with seed 7321 and takes the reported prefixes. For
Electricity, we sort the identifiers of individuals with exactly 12 choice
occasions and take the requested prefix.

\looseness=-1
Only final log likelihood determines the final-likelihood agreement column.
The parameter and prediction differences reported below provide additional
output diagnostics. \textit{Fail} denotes an attempted run that ends
unsuccessfully, and \textit{Timeout} denotes a run that reaches the 300-second
limit.

\subsection{Latent-class estimation}

\looseness=-1
We evaluate latent-class estimation on a controlled data-generating process and
on Swissmetro data processed with the main benchmark's filtering, scaling, and
availability rules. The model has two class-specific MNL kernels and one
observed class-membership covariate \(q_n\). The synthetic cases draw \(q_n\)
from a standard normal distribution, while the Swissmetro cases use the annual
rail-pass indicator. Letting \(\mathcal{S}=\{1,2\}\), the observed-choice
likelihood contribution is
\begin{align}
    L_n
    &=\sum_{s\in\mathcal{S}}\pi_{n,s}P_{n,y_n\mid s},
    &
    \pi_{n,2}
    &=\Lambda\!\left(\delta_0+\delta_z q_n\right),
    \qquad \pi_{n,1}=1-\pi_{n,2}.
    \label{ec:eq:latent_class_replay}
\end{align}
\looseness=-1
The synthetic cases generate choices from the structure in
Eq.~\eqref{ec:eq:latent_class_replay} using seed
\(100+\lvert\mathcal{N}\rvert\). In the Swissmetro
cases, the class-specific utilities use travel time and alternative-specific constants,
while the annual rail-pass indicator enters class membership. The model has
eight free parameters. Each class has one generic travel-time coefficient and
two alternative-specific constants (ASCs), with the ASC for alternative A
normalized to zero. The class-2 membership intercept and rail-pass coefficient
complete the parameter vector. Both data sources use the same starts:
\((\mathrm{ASC}_{B,1},\mathrm{ASC}_{C,1},\beta_{x,1})
=(-0.30,0.15,-0.80)\),
\((\mathrm{ASC}_{B,2},\mathrm{ASC}_{C,2},\beta_{x,2})
=(0.50,-0.20,-0.20)\), and
\((\delta_0,\delta_z)=(-0.10,0.60)\).

\looseness=-1
Latent classes are aligned before comparison by requiring
\(\beta_{x,1}<\beta_{x,2}\). If an estimator returns the reverse ordering, the two
class-specific utility blocks are exchanged and
\((\delta_0,\delta_z)\) is negated so that class 1 remains the membership
reference. The diagnostics report maximum pairwise differences in aligned natural-scale
parameters, mean class shares
\(\bar{\boldsymbol{\pi}}=\lvert\mathcal{N}\rvert^{-1}
\sum_{n\in\mathcal{N}}\boldsymbol{\pi}_n\), and predicted choice
probabilities from each estimator.

\begin{table}[H]
\caption{Latent-class full-estimation runtime comparison (seconds).}
\label{ec:tab:latent_class_likelihood}
\centering
\scriptsize
\setlength{\tabcolsep}{4pt}
\begin{tabular*}{\linewidth}{@{\extracolsep{\fill}}lrrrrrrrl@{}}
\toprule
Data & $\lvert\mathcal{N}\rvert$ & TorchDCM & Biogeme & Apollo &
\(\Delta_{\theta}\) & \(\Delta_{\bar{\pi}}\) & \(\Delta_P\) &
\makecell{Final-likelihood\\agreement?} \\
\midrule
Synthetic & 2,000 & 0.014 & 2.134 & 0.502 & 2.23e-4 & 4.51e-5 & 7.32e-5 & Yes \\
Synthetic & 5,000 & 0.025 & 2.017 & 0.733 & 2.94e-4 & 3.19e-5 & 5.91e-5 & Yes \\
Synthetic & 10,000 & 0.038 & 2.009 & 1.253 & 2.13e-6 & 4.70e-7 & 2.60e-7 & Yes \\
\midrule
Swissmetro & 2,000 & 0.020 & 2.153 & 0.551 & 1.90e-3 & 9.57e-5 & 5.69e-5 & Yes \\
Swissmetro & 3,500 & 0.027 & 2.261 & 0.847 & 1.74e-5 & 2.17e-6 & 3.37e-6 & Yes \\
Swissmetro & 5,000 & 0.032 & 2.035 & 0.985 & 1.04e-3 & 2.20e-5 & 3.44e-5 & Yes \\
\bottomrule
\end{tabular*}
\end{table}

\looseness=-1
All six cases attain final-likelihood agreement after class alignment. Across
the cases, the maximum parameter, mean-class-share, and probability
differences are \(1.90\times10^{-3}\), \(9.57\times10^{-5}\), and
\(7.32\times10^{-5}\). TorchDCM has the lowest runtime in every case,
requiring 0.014--0.038 seconds, compared with 0.502--1.253 seconds for Apollo
and 2.009--2.261 seconds for Biogeme. At these scales, fixed estimation and
covariance overhead account for Biogeme's roughly two-second runtime.

\subsection{Hybrid-choice estimation}

\looseness=-1
The hybrid-choice experiment uses controlled synthetic data and Optima data.
Both specifications contain the same structural equation, continuous
measurement equations, and binary choice kernel. Let \(q_n\) denote the structural
covariate, \(x_n\) the choice covariate, and \(c_n\in\{A,B\}\) the observed
choice. The latent variable is
\(I_n=\gamma q_n+\sigma_I\xi_n\), where
\(\xi_n\sim\mathcal{N}(0,1)\). For the two indicators
\(h\in\mathcal{H}=\{1,2\}\), the Gaussian measurement equation is
\(y_{n,h}=\alpha_h+\lambda_h I_n+\nu_{n,h}\), with
\(\nu_{n,h}\sim\mathcal{N}(0,\sigma_{y_h}^2)\) and density \(f_h\).
The binary choice kernel satisfies
\(P(c_n=B\mid I_n)=
\Lambda(\mathrm{ASC}_B+\beta_x x_n+\beta_I I_n)\), with the complementary
probability assigned to alternative A. The joint likelihood contribution is
approximated by
\begin{align}
    L_n^{\mathrm{HC}}
    &=\frac{1}{R}\sum_{r=1}^{R}
      P\!\left(c_n\mid I_{n,r}\right)
      \prod_{h\in\mathcal{H}}
      f_h\!\left(y_{n,h}\mid I_{n,r}\right),
    &
    I_{n,r}
    &=\gamma q_n+\sigma_I z_r.
    \label{ec:eq:hybrid_replay}
\end{align}
\looseness=-1
In Eq.~\eqref{ec:eq:hybrid_replay}, \(\mathcal{H}\) indexes the two continuous
indicators, and \(z_r\) is the shared standard-normal draw. Identification fixes
the structural intercept
at zero, the first indicator intercept at zero, and its loading at one. These
restrictions fix the latent-variable location, scale, and orientation. The
constant for alternative A is also fixed at zero. The second indicator
intercept and loading are estimated.

The model has nine free parameters with common starting values
\((\mathrm{ASC}_B,\beta_x,\gamma,\sigma_I,\beta_I)
=(0.10,0.40,0.40,0.60,0.60)\) and
\((\sigma_{y_1},\alpha_2,\lambda_2,\sigma_{y_2})
=(0.80,0.10,0.60,0.90)\). We constrain all three standard deviations to be
positive. TorchDCM uses an exponential map for the hybrid-choice standard deviations, Apollo uses
a softplus map, and Biogeme applies a \(10^{-5}\) lower bound. We compare the
resulting estimates on their positive natural scale. The \(R\) draws are one-dimensional
antithetic standard normals. Synthetic data use seed
\(200+\lvert\mathcal{N}\rvert\), and their integration draws use seed
\(1200+\lvert\mathcal{N}\rvert\); Optima integration draws use seed
\(1400+\lvert\mathcal{N}\rvert\). We construct each draw set from
\(\lceil R/2\rceil\) standard-normal values and their negatives. For Optima, we
standardize \(q_n=\texttt{ScaledIncome}\),
\(x_n=\texttt{TimePT\_scaled}-\texttt{TimeCar\_scaled}\), and
\((y_{n,1},y_{n,2})=(\texttt{Envir01},\texttt{Envir02})\) over all 1,298
retained observations before taking the reported prefixes. For
predicted-probability differences, we condition on the observed indicators and
use the common draws.

\begin{table}[htb]
\caption{Hybrid-choice full-estimation runtime comparison (seconds).}
\label{ec:tab:hybrid_likelihood}
\centering
\scriptsize
\setlength{\tabcolsep}{4pt}
\begin{tabular*}{\linewidth}{@{\extracolsep{\fill}}lrrrrrrrl@{}}
\toprule
Data & $\lvert\mathcal{N}\rvert$ & $R$ & TorchDCM & Biogeme & Apollo &
\(\Delta_{\theta}\) & \(\Delta_P\) &
\makecell{Final-likelihood\\agreement?} \\
\midrule
Synthetic & 500 & 32 & 0.049 & 39.313 & 0.932 & 2.28e-6 & 9.91e-7 & Yes \\
Synthetic & 2,000 & 64 & 0.281 & 163.670 & 5.937 & 3.96e-6 & 3.12e-6 & Yes \\
Synthetic & 10,000 & 128 & 3.505 & Timeout & 71.886 & 2.39e-7 & 6.28e-8 & Yes \\
\midrule
Optima & 500 & 32 & 0.055 & 39.520 & 1.205 & 2.89e-6 & 5.27e-7 & Yes \\
Optima & 1,000 & 64 & 0.129 & 162.814 & 3.718 & 1.21e-5 & 3.56e-6 & Yes \\
Optima & 1,298 & 128 & 0.296 & Timeout & 9.203 & 5.13e-6 & 1.24e-6 & Yes \\
\bottomrule
\end{tabular*}
\end{table}

\looseness=-1
All six cases attain final-likelihood agreement. The maximum parameter and
conditional choice-probability differences are \(1.21\times10^{-5}\) and
\(3.56\times10^{-6}\). Four cases provide three-way comparisons. For the two
128-draw cases, we compare TorchDCM with Apollo because Biogeme reaches the time
limit. Runtime increases with both \(\lvert\mathcal{N}\rvert\) and \(R\).
TorchDCM requires 0.049--3.505 seconds and Apollo 0.932--71.886 seconds.
Biogeme requires about 39 seconds with 32 draws and about 163 seconds with 64
draws.

\subsection{Panel estimation}

\looseness=-1
The panel experiment uses controlled repeated-choice data and the Electricity
panel. Let
\(\mathcal{N}\) be the set of individuals, \(\mathcal{P}_n\) the set of
observed choice occasions for individual \(n\), and \(n_u\) the observation at
occasion \(u\in\mathcal{P}_n\). The complete set of choice observations is
\(\mathcal{O}=\{n_u:n\in\mathcal{N},\,u\in\mathcal{P}_n\}\). The synthetic
data-generating process assigns one random coefficient
to each individual, while each retained Electricity individual contributes 12
choice occasions. For
choice occasion \(u\in\mathcal{P}_n\), alternative \(j\), and draw \(r\), the
utility is
\begin{align}
    V_{n_u,j,r}
    &=\mathrm{ASC}_j+
      \left(\beta_x+\sigma_x z_r\right)x_{n_u,j},
    &
    z_r&\sim\mathcal{N}(0,1),
    \qquad \mathrm{ASC}_{A}=0.
    \label{ec:eq:panel_utility}
\end{align}
\looseness=-1
In Eq.~\eqref{ec:eq:panel_utility}, the same random coefficient applies across
all occasions for an individual. Each synthetic alternative-specific attribute is
drawn from a standard normal distribution. For every Electricity case, we sort
the identifiers of individuals with exactly 12 occasions and retain the
required prefix. The
four price columns are pooled and standardized with one mean and standard
deviation computed within that subset. Neither specification includes a
separate panel-level observed covariate.

\looseness=-1
Both datasets use five free parameters with common starts
\((\mathrm{ASC}_B,\mathrm{ASC}_C,\mathrm{ASC}_D,\beta_x,\sigma_x)
=(0.20,-0.10,0.05,-0.50,0.40)\). We estimate and compare \(\sigma_x\) on its
positive natural scale, as in the hybrid-choice experiment. The
draws are one-dimensional antithetic standard normals. Synthetic data use seed
\(500+\lvert\mathcal{N}\rvert+\lvert\mathcal{P}_n\rvert\), and their
integration draws use seed \(1500+\lvert\mathcal{N}\rvert+
\lvert\mathcal{P}_n\rvert\), where \(\lvert\mathcal{N}\rvert\) counts
individuals in these expressions. Electricity integration draws use
\(1800+\lvert\mathcal{N}\rvert\). The \texttt{person\_id} field maps each ordered
\texttt{obs\_id} to its individual. In both settings,
\(\mathcal{P}_n\) contains individual \(n\)'s repeated choice occasions. We
therefore multiply conditional probabilities across each individual's
occasions before averaging over draws:
\begin{align}
    \ell
    &=\sum_{n\in\mathcal{N}}
      \log\!\left[
      \frac{1}{R}\sum_{r=1}^{R}
      \prod_{u\in\mathcal{P}_n}P_{n_u,y_{n_u}\mid r}
      \right].
    \label{ec:eq:panel_replay}
\end{align}
\looseness=-1
In Eq.~\eqref{ec:eq:panel_replay}, \(P_{n_u,y_{n_u}\mid r}\) is the draw-\(r\)
probability assigned to the observed alternative \(y_{n_u}\) in observation
\(n_u\).
The prediction diagnostic compares each individual's fitted probability of
the observed choice sequence. Across the synthetic cases, we increase the
numbers of individuals, repeated choices, and draws. Across the Electricity
cases, we increase the number of complete individuals and draws while holding
\(\lvert\mathcal{P}_n\rvert=12\).

\begin{table}[htb]
\caption{Panel full-estimation runtime comparison (seconds).}
\label{ec:tab:panel_likelihood}
\centering
\scriptsize
\setlength{\tabcolsep}{1.5pt}
\begin{tabular*}{\linewidth}{@{\extracolsep{\fill}}lrrrrrrrrrl@{}}
\toprule
Data & $\lvert\mathcal{N}\rvert$ & $\lvert\mathcal{P}_n\rvert$ & $\lvert\mathcal{O}\rvert$ & $R$ &
TorchDCM & Biogeme & Apollo & \(\Delta_{\theta}\) & \(\Delta_{P_n}\) &
\makecell{Final-likelihood\\agreement?} \\
\midrule
Synthetic & 250 & 2 & 500 & 32 & 0.024 & 29.843 & 0.539 & 6.57e-6 & 3.18e-6 & Yes \\
Synthetic & 500 & 4 & 2,000 & 64 & 0.101 & 157.266 & 3.168 & 1.96e-7 & 2.87e-8 & Yes \\
Synthetic & 1,250 & 8 & 10,000 & 128 & 2.557 & Timeout & 32.145 & 4.56e-7 & 3.52e-8 & Yes \\
\midrule
Electricity & 100 & 12 & 1,200 & 32 & 0.038 & Timeout & 0.967 & 3.88e-7 & 2.36e-11 & Yes \\
Electricity & 250 & 12 & 3,000 & 64 & 0.147 & Timeout & 4.710 & 8.60e-6 & 4.48e-10 & Yes \\
Electricity & 348 & 12 & 4,176 & 128 & 0.597 & Timeout & 18.058 & 4.29e-7 & 6.09e-11 & Yes \\
\bottomrule
\end{tabular*}
\end{table}

\looseness=-1
All six panel cases attain final-likelihood agreement. The maximum parameter
and individual sequence-probability differences are
\(8.60\times10^{-6}\) and \(3.18\times10^{-6}\). The two smaller synthetic
cases provide three-way comparisons. For the remaining four, we compare TorchDCM with
Apollo because Biogeme reaches the time limit. TorchDCM requires
0.024--2.557 seconds across the synthetic cases, compared with
0.539--32.145 seconds for Apollo. Biogeme finishes the first two cases in
29.843 and 157.266 seconds. For Electricity, TorchDCM requires
\(0.038\)--\(0.597\) seconds and Apollo \(0.967\)--\(18.058\) seconds. Longer
choice histories and more draws increase panel-estimation time.

The unified result interface carries each fitted model through inference,
convergence diagnostics, and reproducible export. The ordered-response study
establishes full-output agreement in likelihoods, natural-scale parameters, and
predictions for all 108 actual-data cases and 20 synthetic model--case
combinations. The latent-class, hybrid-choice, and panel studies likewise show
final-likelihood agreement and small aligned parameter and prediction
differences whenever at least two comparable estimators remain. Together, these results
extend the main-text evidence for model coverage and numerical agreement while
showing that the supported model families use a unified estimation and reporting
pipeline.

\end{document}